\documentclass{article}

\usepackage{PRIMEarxiv}

\usepackage[utf8]{inputenc} 
\usepackage[T1]{fontenc}    
\usepackage{hyperref}       
\usepackage{url}            
\usepackage{booktabs}       
\usepackage{amsfonts}       
\usepackage{nicefrac}       
\usepackage{microtype}      
\usepackage{lipsum}
\usepackage{fancyhdr}       
\usepackage{graphicx}       
\graphicspath{{media/}}     

\newif\ifsubmit\submittrue

\newcommand\mf[1]{\ifsubmit\else {\color{red}  [MF: #1]}\fi}

\newcommand\mfhide[1]{}

\newcommand\code[1]{{\small\texttt{#1}}}
\newcommand\js{JavaScript}
\newcommand\clutch{\textsc{Clutch}}
\newcommand\clite{\textsc{ClutchLite}}
\newcommand\fz{Fuzzilli}
\newcommand\fzc{FClutch}
\newcommand\fj{FuzzJIT}
\newcommand\fjc{FJClutch}
\newcommand\jp{JIT-Picker}
\newcommand\jpc{JPClutch}

\newcommand\jpa{JPClutchLite}
\newcommand\fja{FJClutchLite}
\newcommand\fza{FClutchLite}

\newcommand{\eg}{{\it e.g.}}%
\newcommand{\ie}{{\it i.e.}}%
\newcommand{\etal}{{\it et al.}}%

\usepackage{newfloat}
\usepackage{float}
\usepackage{listings}
\usepackage{stackengine}

\floatstyle{ruled}
\newfloat{listing}{tb}{lst}{}
\floatname{listing}{Listing}

\usepackage{graphicx}
\usepackage{textcomp}
\usepackage{enumitem}
\usepackage{booktabs}
\usepackage{multirow}
\usepackage{pifont}
\usepackage{xcolor, colortbl}
\usepackage{color}
\usepackage{tikz}
\usepackage{adjustbox}
\usepackage{bm}
\usepackage{array}

\usepackage{hyperref}
\newcolumntype{C}[1]{>{\centering\arraybackslash}p{#1}}

\newcommand\YAMLcolonstyle{\color{red}\mdseries\footnotesize}
\newcommand\YAMLkeystyle{\color{black}\bfseries\footnotesize}
\newcommand\YAMLvaluestyle{\color{blue}\mdseries\footnotesize}

\makeatletter

\newcommand\language@yaml{yaml}

\expandafter\expandafter\expandafter\lstdefinelanguage
\expandafter{\language@yaml}
{
  keywords={true,false,null,y,n},
  keywordstyle=\color{darkgray}\bfseries\scriptsize,
  basicstyle=\YAMLkeystyle\ttfamily\linespread{0.9}\scriptsize,                                 
  sensitive=false,
  comment=[l]{\#},
  morecomment=[s]{/*}{*/},
  commentstyle=\color{purple}\ttfamily\scriptsize,
  stringstyle=\YAMLvaluestyle\ttfamily\scriptsize,
  moredelim=[l][\color{orange}]{\&},
  moredelim=[l][\color{magenta}]{*},
  moredelim=**[il][\YAMLcolonstyle{:}\YAMLvaluestyle\scriptsize]{:},   
  morestring=[b]',
  morestring=[b]",
  literate =    {---}{{\ProcessThreeDashes}}3
                {>}{{\textcolor{red}\textgreater}}1     
                {|}{{\textcolor{red}\textbar}}1 
                {\ -\ }{{\mdseries\ -\ }}3,
}
\colorlet{punct}{red!60!black}
\definecolor{delim}{RGB}{20,105,176}
\colorlet{numb}{magenta!60!black}

\definecolor{lightgray}{rgb}{.9,.9,.9}
\definecolor{darkgray}{rgb}{.4,.4,.4}
\definecolor{purple}{rgb}{0.65, 0.12, 0.82}

\lstdefinelanguage{JavaScript}{
  keywords={typeof, new, true, false, catch, function, return, null, catch, switch, var, if, in, while, do, else, case, break},
  basicstyle=\small\ttfamily,
  keywordstyle=\color{blue}\bfseries,
  ndkeywords={class, export, boolean, throw, implements, import, this},
  ndkeywordstyle=\color{darkgray}\bfseries,
  identifierstyle=\color{black},
  sensitive=false,
  comment=[l]{//},
  morecomment=[s]{/*}{*/},
  commentstyle=\color{purple}\ttfamily,
  stringstyle=\color{red}\ttfamily,
  morestring=[b]',
  morestring=[b]",
}

\lstdefinelanguage{json}{
    basicstyle=\small\ttfamily,
    numbers=left,
    stepnumber=1,
    showstringspaces=false,
    breaklines=true,
    literate=
     *{0}{{{\color{numb}0}}}{1}
      {4}{{{\color{numb}4}}}{1}
      {5}{{{\color{numb}5}}}{1}
      {6}{{{\color{numb}6}}}{1}
      {7}{{{\color{numb}7}}}{1}
      {8}{{{\color{numb}8}}}{1}
      {9}{{{\color{numb}9}}}{1}
      {:}{{{\color{punct}{:}}}}{1}
      {,}{{{\color{punct}{,}}}}{1}
      {\{}{{{\color{delim}{\{}}}}{1}
      {\}}{{{\color{delim}{\}}}}}{1}
      {[}{{{\color{delim}{[}}}}{1}
      {]}{{{\color{delim}{]}}}}{1},
}

\lst@AddToHook{EveryLine}{\ifx\lst@language\language@yaml\YAMLkeystyle\fi}
\makeatother

\newcommand\ProcessThreeDashes{\llap{\color{cyan}\mdseries-{-}-}}

\usepackage{amsmath}
\usepackage{amssymb}
\usepackage{mathtools}
\usepackage{algorithm}
\usepackage{algorithmic}
\usepackage{amsthm}
\usepackage{subfigure}

\title{\clutch\ Control: An Attention-based Combinatorial Bandit for Efficient Mutation in \js\ Engine Fuzzing
}

\author{
  Myles Foley\textsuperscript{1}, Sergio Maffeis\textsuperscript{1}, Muhammad Fakhrur Rozi\textsuperscript{2}, Takeshi Takahashi\textsuperscript{2} \\ \\
  \textsuperscript{1}Imperial College London \\
  \textsuperscript{2}National Institute of Information and Communications Technology 
}

\begin{document}
\maketitle

\begin{abstract}
\js\ engines are widely used in web browsers, PDF readers, and server-side applications. 
The rise in concern over their security has led to the development of several targeted fuzzing techniques.
However, existing approaches use random selection to determine where to perform mutations in \js\ code.
We postulate that the problem of selecting better mutation targets is suitable for combinatorial bandits with a volatile number of arms. Thus, we propose \clutch, a novel deep combinatorial bandit that can observe variable length \js\ test case representations, using an attention mechanism from deep learning.
Furthermore, using Concrete Dropout, \clutch\ can dynamically adapt its exploration.
We show that \clutch\ increases efficiency in \js\ fuzzing compared to three state-of-the-art solutions by increasing the number of valid test cases and coverage-per-testcase by, respectively, 20.3\% and 8.9\% on average.
%
%
In volatile and combinatorial settings we show that \clutch\ outperforms state-of-the-art bandits, achieving at least 78.1\% and 4.1\% less regret in volatile and combinatorial settings, respectively.

\end{abstract}


 \section{Introduction}
\js\ is commonly used across the web, with a consistent year-on-year increase~\cite{http_archive_et_al_web_2023,http_archive_et_al_web_2022}. 
This prevalence is in part due to frameworks that allow for easy and lightweight deployment such as jQuery, React, and core-js. 
\js\ is a dynamically typed, prototype based, object oriented language, that allows for flexibility of types and properties during runtime as the code executes.
As such, it is used for a variety of tasks in modern web sites, including animations, logging, content loading, API calls. 
Due to the large volume of \js\ on the web, and its varied functionality, \js\ engines use Just-In-Time (JIT) compilers. 
These JIT compilers have become a focus of security concerns, due to the rising number of vulnerabilities they contain~\cite{park_fuzzing_2020}.

Fuzzing is one of the most common and efficient methodologies to find vulnerabilities in software~\cite{zalewski_american_2007}. 
Broadly, these approaches generate new \js\ and then mutate test cases to change the structure, syntax, or behaviour of the test case, this is in the hope of triggering new paths in the \js\ engine and find potential vulnerabilities or bugs.
Fuzzing of \js\ engines has spawned different methodologies 
ranging from neural networks for code generation~\cite{lee_montage_2020} to heuristics ~\cite{park_fuzzing_2020}.
%
However, while they focus on developing new mutations, testing heuristics, or oracles, they apply mutations at random locations within test cases.
This behaviour reduces the number of successful test cases, and leads to inefficient search of system under test.
%
%

\begin{lstlisting}[language=javascript, caption={Javascript test case that always throws an error (Line 6) without clearing the cache leading to a segmentation fault in  SpiderMonkey\protect\footnotemark.\vspace{+1mm}}, label={lst:jsc_bug example},frame=tb,float=tp,floatplacement=tbp]
for (let v3 = 0; v3 < 100; v3++) {
    function v4(v5,v6) {}
    function v7(v8,v9) {
        v14 = BigInt(-1.7976931348623157e+308);
        try {
            v15 = BigInt.asUintN(639625512,v14);  
        } catch(v17) {}
    }
    v18 = v4.prototype;
    v18.b = v7;
    v19 = new v4();
    v20 = v19.b();
}
\end{lstlisting}
\footnotetext{\url{https://bugzilla.mozilla.org/show_bug.cgi?id=1745907}\label{fnlabel}}

%
Consider the test case in Listing~\ref{lst:jsc_bug example} that causes a bug due to the following behaviour. 
%
%
On line 4 \code{v14} is assigned as a \code{BigInt}, and on line 6 a call to \code{asUintN} always throws an error as the result is too large.
The optimiser then does not assign the result to \code{v15}, leaving it in the cache.
Finally, due to the repeated calls from the \code{for} loop a segmentation fault occurs in FireFox's Spidermonkey engine\footref{fnlabel}.
%
However, if the variable \code{v14} on line 4 was mutated to a different type instead, e.g. \code{Int} the crash would not have been triggered.

We postulate it is possible to \emph{learn} where to perform such a mutation, instead of random selection.
The problem of selecting a given location in a \js\ program to mutate is similar to that of contextual Combinatorial Bandits (CBs). 
At each timestep a representation of a test case can be observed (the contexts of arms) and locations are selected (selection of a subset of arms).
The test case is immediately run and feed and feedback on the performance of the test case is retrieved (reward is received based on select arms). 
However, to apply a CB approach to such a task is challenging. 
%
%
First, the CBs must generalise quickly to the large state-action space of potential test cases generated:
\js\ fundamentals are defined in over 800 pages of technical detail in the 15th ECMA standards~\cite{ecma_international_ecmascript_2024}.
%
Second, the arms of the CB are \textit{volatile}: \js\ test cases are of variable size and structure.
%
Third, the exploration of arms must be \emph{dynamic}: random selection of testcase locations should change during testing of the \js\ engines.

To overcome these challenges, we develop \clutch, a novel deep CB.
Given a representation of a \js\ test case, and a mutation to apply, \clutch\ will select where to apply the mutation, immediately receiving feedback from test case performance that is used to guide learning via a reward. 
\clutch\ uses an attention-based neural network with concrete dropout, allowing it handle different sizes of \js\ test case, and adapt its exploration as fuzzing progresses.
%
We demonstrate the utility of \clutch\ in a two stage evaluation. First, we embed \clutch\ inside three state-of-the-art \js\ fuzzers, comparing approaches focusing on their \emph{efficiency} over the number of test cases they are able to produce.
%
%
Second, to compare with state-of-the-art bandits in terms of regret we evaluate in separate volatile and combinatorial settings where ground truth rewards are known. 
%
In summary, our contributions are:
\begin{itemize}
    \item  We develop \clutch, a novel attention-based deep combinatorial bandit which overcomes challenges of volatility and dynamic control of exploration during software testing, by using the attention mechanism of Pointer Networks, and Bayesian exploration via Concrete Dropout.
    
    \item We demonstrate that \clutch\ is able to learn \emph{where} to perform mutations in test cases for fuzzing \js\ engines.

    \item Through a thorough evaluation on different \js\ engines we show \clutch\ increases the efficiency of testing compared to state-of-the-art approaches. 
    %
    \item We show that the design of \clutch\ outperforms state-of-the-art bandits in bandit-based evaluations to achieve lower regret. Underpinning the potential impact in other domains beyond software engineering.
    
\end{itemize}

\section{Background}

\paragraph{Contextual Bandit Variants.} 

Deep Neural Networks (DNNs) have been applied to contextual bandit problems. One of the first was NeuralUCB, which used a DNN to approximate the reward in the Upper Confidence Bound (UCB) algorithm~\cite{zhou_neural_2020}. This was closely followed the use of DNNs for Thompson Sampling (TS)~\cite{zhang_neural_2021}, which introduces a posterior over the neural network to approximate the reward instead of direct approximation.
%
Contextual CBs are a special class of bandit where a bandit selects multiple arms at each round. The CB problem is closely related to the one considered in this paper. 
Neural CBs were proposed by Hwang \etal~\cite{hwang_combinatorial_2023}, and are applied to both UCB (CN-UCB) and TS (CN-TS) settings. Both CN-UCB and CN-TS sample directly from the reward predicted by a DNN to select the arms to play.
Volatile CBs (VCBs) consider the case where the number of arms can change with the task that is being conducted.
To the best of our knowledge, the DNNs haven't been applied to volatile bandit conditions due to the variable number of arms.
The first work to consider VCB problems was CC-MAB~\cite{nika_contextual_2020} an online policy that is formed by creating identical hypercubes that exist within the space of possible arms. A similar approach is taken by ACC-UCB~\cite{nika_contextual_2020}, that adaptively discretises the hypercubes to estimate outcomes of different arms. As a result, ACC-UCB functions only in a two-dimensional feature space of arms.








\paragraph{Fuzzing Strategies.} 
The goal of fuzzing \js\ engines is to trigger abnormal behaviour, corresponding to a bug or vulnerability. 
%
Fuzzing tools either tend to generate new test cases or mutate existing ones, at random or following some predefined heuristics.
Code coverage is a common metric to evaluate the performance of the mutated test cases, providing a practical estimate of the portion of the search space explored, such fuzzers are referred to as \emph{greybox}.
%
%
However, only a small fraction of test cases will find bugs in the test program, leading to a high number of executed tests~\cite{bohme_coverage-based_2016}.
Hence, a key objective for practical fuzzing solutions is to increase the efficiency of bug finding by reducing the volume of test cases that do not result in new behaviour or do not pass input validation~\cite{wang_systematic_2020}.

There has been no prior work in fuzzing to \emph{learn} where to perform mutations in a test case. 
Limpanukorn \etal~\cite{limpanukorn_fuzzing_2024} use a tree-like representation of a test case to  deterministically perform mutations at locations dependent on shared ancestor and sibling nodes.
However, this approach is developed for low-level language representations, and not for \js.
%
Orthogonal works have focused on different aspects of fuzzing. Weissberg \etal~\cite{weissberg_sok_2024} examined different methods to select areas in the \emph{target} to fuzz in directed fuzzing~\cite{lyu_mopt_2019,assiri_software_2024,srivastava_one_2022}. 
%
%
Zhao \etal~\cite{zhao_alphuzz_2022} investigated the scheduling of seeds using Monte-Carlo Tree Search approach.
Lemieux \etal~\cite{eom_fuzzing_2024} mask locations at random, then use a language model to generate a mutation. The model is then fine-tuned via reinforcement learning to improve testing.
While these works investigate selection of target code, scheduling mutators, or scheduling seeds, \clutch\ learns to select where in the \emph{test case} to perform mutations in greybox fuzzing.

More recent research has focused specifically on the JIT compiler to target the deeper functionality and harder-to-find vulnerabilities contained within. 
Several of these approaches are based on Fuzzilli~\cite{gros_fuzzilli_2023}, which provided modular and customisable templates for generating \js\ for particular engines; in addition to mutations designed to trigger JIT compilation.
%
%
%
OptFuzz~\cite{wang_optfuzz_2024}, builds on the \fz\ architecture to use optimisation paths as a performance metric.
The authors of OptFuzz use an approximation of the optimisation path coverage as feedback to guide seed scheduling and preservation.


\paragraph{RL for Testing.}

RL techniques have been developed to automate many different types of testing. 
One of the first works to investigate RL for automating fuzzing was from B\"ottinger \etal~\cite{bottinger_deep_2018} which focused on fuzzing C compliers. 
This has lead to many works that use RL for fuzzing different compilers \cite{li_alphaprog_2022,li_fuzzboost_2022,sablotny_reinforcement_2023}. 
Zheng \etal~\cite{zheng_automatic_2021} used an RL agent with an intrinsic reward function to increase coverage in web crawling.
While Tsingenopoulos \etal~\cite{tsingenopoulos_captcha_2022}, used an alternative deep RL model to bypass captchas in web applications. 
RL has been used for a number of different security focused testing applications, including SQL injection~\cite{erdodi_simulating_2021,wahaibi_sqirl_2023}, cross site scripting~\cite{lee_link_2022,foley_haxss_2022}, denial-of-service~\cite{mcfadden_wendigo_2024}, and cache timing attacks~\cite{luo_autocat_2023}.


\paragraph{Bandits for Software Testing.}

Bandits have been used for software testing for seed scheduling.  Yue \etal~\cite{yue_ecofuzz_2020} propose a variant of an adversarial Multi-Armed Bandit (MAB), which models seed selection using the estimated reward probability.
Wang \etal~\cite{wang_reinforcement_2021} use a similar problem formulation using an MAB, they further consider feedback in the form of hierarchical feedback on coverage of test cases. 
More recently Luo \etal~\cite{luo_make_2024} use Thompson Sampling (TS) for seed selection, with a sparse reward function.
McFadden \etal~\cite{mcfadden2025drmddeepreinforcementlearning} use PPO as a one-step MDP to classify malware samples additionally combining active learning, rejection.

\section{Fuzzing JavaScript Engines.}


Fuzzing has become the de facto approach to find bugs and vulnerabilities in \js\ engines, leading to a number of works on this.
LangFuzz ~\cite{christian_fuzzing_2012}, creates an Abstract Syntax Tree (AST) of \js\ functions, selecting nodes to mutate at random.
IFuzzer~\cite{veggalam_ifuzzer_2016} makes use of a similar strategy, but improves effectiveness with genetic programming, and a fitness function from greybox feedback on the test case.
%
Montage~\cite{lee_montage_2020} uses Natural Language Processing (NLP) to generate JavaScript subtrees at random locations in a test case's AST.
%
DIE~\cite{park_fuzzing_2020} is a mutation based fuzzer that aims to preserve type and structural properties of test cases.
CodeAlchemist~\cite{han_codealchemist_2019}, reduces test cases to fragments, using static and dynamic analysis to infer types, and then reassemble test cases. 
In both Superion~\cite{wang_superion_2019}, and SkyFire~\cite{wang_skyfire_2017}, the authors use grammar based methods for mutation and generation, respectively. 
Nautilus~\cite{aschermann_nautilus_2019} takes this a stage further: generating its own test cases and then mutating their ASTs.
While Salls \etal~\cite{salls_token-level_2021} use a `token level' approach to perform mutations at a higher level, to avoid using grammars.

Our approach, \clutch, instead uses a deep CB to select the locations to mutate. 
\clutch\ is independent from the JavaScript engine, and the fuzzer. 
We will place \clutch\ inside three state-of-the-art fuzzers for evaluation in Section~\ref{eval:ftc}. 
Each fuzzer approaches the problem differently, yet share similarities, we briefly describe the salient points of these fuzzers related to \clutch\ here.  

\fz~\cite{gros_fuzzilli_2023} is a greybox fuzzer targeting crashes in the JIT compiler of JS engines. 
\fz\ leverages coverage feedback and semantic correctness preservation in order to improve its ability to fuzz. 
%
%
Several approaches are based on \fz, due to the modular and customisable templates for generating \js\ for particular engines.

\fj~\cite{wang_fuzzjit_nodate} is a tool based on the \fz\ architecture, but develops JIT-specific oracles and triggers. Specifically, \fj\ wraps each generated function in a wrapper that can be called repeatedly to trigger JIT compilation. The wrapped output can then be compared before and after JIT compilation to determine errors that result form the JIT. 
\fj\ implements its oracle to perform type-specific comparisons, using `deep equality' to find if there are differences in objects, arrays, or individual values, and in such cases reporting an error.



\jp~\cite{bernhard_jit-picking_2022} uses differential testing on the backbone of \fz\ to detect subtle non-crash bugs. 
It uses both the interpreter and the JIT compiler as separate bug oracles, using the fact that both demand the interpreted and compiled code to perform computations in a strict and substitutable way.
This abstracts away from implementation-dependent behaviour, allowing \jp\ to perform differential testing in individual engines without need to instrument the \js\ code being tested.




%

\setbox0=\hbox{%
\begin{minipage}{0.45\textwidth}
\begin{lstlisting}[
language=javascript,
basicstyle={\tiny\ttfamily},
identifierstyle={\color{black}},
tabsize=2,
numbersep=8pt,
caption={\js\ source code of a test case.\vspace{+1mm}}, label={lst:js_example},frame=tb,escapeinside={(*}{*)},
morekeywords ={class,run,const}
]
const v2 = new Uint32Array(256);
const v3 = [0, 1, 2, 3, 4, 5, 6];

for (const v4 of v2) {
    ~v3[v4];
}

(*\color{white}*)
\end{lstlisting}
\end{minipage}
}
\savestack{\listingA}{\box0}

\setbox0=\hbox{%
\begin{minipage}{0.45\textwidth}
\begin{lstlisting}[
language=javascript,
basicstyle={\tiny\ttfamily},
identifierstyle={\color{black}},
tabsize=2,
numbersep=8pt,
caption={\fz\ IR of the code in Listing~\ref{lst:js_example}.\vspace{+1mm}}, label={lst:ir_example},frame=tb,
morekeywords ={LoadInteger,LoadBuiltin,Construct,CreateFloatArray,BeginForOfLoop,GetComputedProperty,UnaryOperation,EndForOfLoop}
]
v0 <- LoadInteger '256'
v1 <- LoadBuiltin 'Uint32Array'
v2 <- Construct v1, [v0]
v3 <- CreateFloatArray [0, 1, 2, 3, 4, 5, 6];
BeginForOfLoop v2 -> v4
    v5 <- GetComputedProperty v3, v4
    v6 <- UnaryOperation '~', v5
EndForOfLoop
\end{lstlisting}
\end{minipage}
}
\savestack{\listingB}{\box0}

\begin{figure*}
\begin{tabular}{cc}
\stackinset{l}{+2cm}{t}{0.15\textwidth}{}{\listingA} & 
\stackinset{l}{+2cm}{t}{0.15\textwidth}{}{\listingB} \\
\end{tabular}
\end{figure*}



%

\paragraph{\js\ representation.}
The use of an \emph{Intermediate Representation (IR)} for \js\ was first introduced by \fz.
%
The IR is based on the how the JIT compiler interprets \js, an example can be seen in Listings~\ref{lst:js_example} and~\ref{lst:ir_example}. 
By performing mutations on the IR of the JIT compiler, instead of the JavaScript source code the chance of `semantically meaningless mutations' can be reduced.  

\paragraph{Mutations.}

Fuzzers we consider are able to generate their own test cases, which in a second stage of testing are then mutated at the IR level.
Key mutations, originally introduced by \fz\ include:

%

\begin{itemize}[nosep]
    \item \code{InputMutator}: replace one input instruction with another. For example, changing \code{v4} on line 6 of  Listing~\ref{lst:ir_example} to \code{v0}.
    
    \item \code{InputMutator} (type aware): replace one input instruction with another, ensuring type preservation.
    
    \item \code{OperationMutator}: mutate parameters of an Operation (\eg\ \code{256} on line 1 of Listing~\ref{lst:ir_example} could be replaced with \code{64}). 

    \item \code{CombineMutator}: insert part of a program into another, renaming variables to avoid collisions/syntax errors.
    
    \item \code{CodeGenMutator}: generate new, random code, placing this in the program, renaming variables to avoid collisions/syntax errors.

    \item \code{SpliceMutator}: copy a self contained part of a program within the corpus, placing this into another to combine their features.
\end{itemize}

The fuzzers we consider apply these mutations on IR instructions \emph{at random} in test cases, making them ideal candidates for prioritisation by \clutch.

\subsection{Challenges in targeting mutations}\label{fuzz:challenge}


Mutational \js\ engine fuzzers traditionally focus on developing new ways to perform mutations~\cite{lee_montage_2020,wang_superion_2019,christian_fuzzing_2012,park_fuzzing_2020}, or new oracles to detect bugs and vulnerabilities~\cite{gros_fuzzilli_2023,bernhard_jit-picking_2022,wang_optfuzz_2024}. 
%
%
%
%
Mutations introduce diversity in the testing set, which may increase coverage, and find bugs. 
In comparison, the problem of \emph{where} to apply these mutations has been overlooked, and they are typically applied at random locations. 
This may be in part due to the difficulty of disentangling the effects on performance of the applied mutation operator and the chosen location.
%
%
We detail below two key challenges in learning where to perform mutations.

\paragraph{Large search space.}
First, \js\ is a highly structured and dynamically typed programming language.
This means that a variable can have arbitrary types dependent on the point of execution. 
Thus, the combinations of possible programs that can be created in \js\ constitutes an enormous search space. 

Mutation based fuzzers, attempt to reduce to problem of creating test cases for fuzzing \js\ engines by applying mutations to existing valid \js.
Conversely some generational based fuzzers use a constrained set of rules, or grammars, to increase the proportion of valid samples. 
Yet, despite this, the search space of test cases that cover new paths or trigger bugs remains vanishingly small compared to the space of syntactically valid \js~\cite{han_codealchemist_2019}. 
This can be considered a challenging search problem: how to mutate the test cases to find the `interesting' test cases that trigger new areas, or behaviours of the \js\ engines.



\paragraph{Complexity of \js.}
Fuzzers traditionally focus on throughput, trying to generate as many test cases as possible. 
They try to do this with some understanding of the `sensible' behaviours that could lead to bugs.
Such behaviours are generally understood by software testers and security practitioners, and used to form the basis of mutations.
In performing these mutations, fuzzers generate a diverse range of \js\ to stress an engine.

However, choosing where to make these mutations makes the problem significantly harder.
Consider again Listing~\ref{lst:jsc_bug example}:  had a mutation on Line 4 caused variable \code{v14} to be of type \code{int}, the bug would not have occurred. 
We can retrospectively see that this was the `best place' for the mutation by analysing the behaviour of the test case.
%
%
Such `post-hoc' analysis is not possible during fuzzing, as the mutation has been applied, and the test case executed.

Finding the `best place' (or at least a `good place') for a mutation requires identifying locations, assigning them preference, and resolving conflicting candidates. 
The design and implementation of appropriate heuristics would require an expansive knowledge of \js and its 800 pages of technical detail as defined in 15th ECMA standards~\cite{ecma_international_ecmascript_2024}.

\section{Our Approach}\label{clutch}

In this Section, we motivate the use of bandits for testing \js\ engines, and then introduce \clutch, a novel deep CB that learns the effective location to mutate \js\ code to increase fuzzing efficiency.
%
%


\subsection{Motivation}

As discussed in Section~\ref{fuzz:challenge}, selecting an effective location to mutate is a challenging task involving covering a large search space, and a deep understanding \js. 
Given these challenges, we believe a Multi-Armed Bandit (MAB) can learn to select a location in a test case, and then immediately receive feedback in the form of coverage.
Framing the problem as a MAB is also advantageous for the computational complexity of fuzzing, as we need to call the bandit only once per test case, reducing overhead.
%
%

%
%
%
%
%

In the traditional MAB actions (selecting an arm) and observations do not require any kind of association. This means that in stationary problems the bandit learns the single `best action', and in non-stationary tasks it learns the `best action' over time \cite{sutton_reinforcement_2018}. 
%
However, test cases have different structures and, as in the example in Listing~\ref{lst:jsc_bug example}, will have different optimal locations to perform a given mutation.
Thus we can consider each test case as a separate task, to optimise for coverage, bugs, or testing deeper into the \js\ engine. 
As we fuzz, generated test cases will be included in the corpus increasing the number of tasks the agent has to learn.
%
Therefore, it is important that we provide a \emph{context} of the current test case to the bandit, enabling the association of actions to observations.
As a result we need a \emph{contextual} bandit that understands the subtleties of the \js\ test cases.

Contextual bandits learn to select one of a fixed number of arms, given a the context of the arm.
This is beneficial as the bandit can observe the effect of each of the arm selected, and detect the shift between different tasks.
%
Yet, the nature of fuzzing deviates from this format, as multiple mutations can be applied simultaneously, making the problem \emph{combinatorial}.
Additionally, the \js\ test cases the bandit sees may be of different structure \emph{and} size, thus the number of locations (arms) is \emph{volatile}. 
%
The challenge is to rapidly adapt to the number of arms at each timestep, while reducing the uncertainty in the location we select.

%




\subsection{\clutch\ Control}

\begin{figure*}
    \centering
    \includegraphics[width=0.7\linewidth]{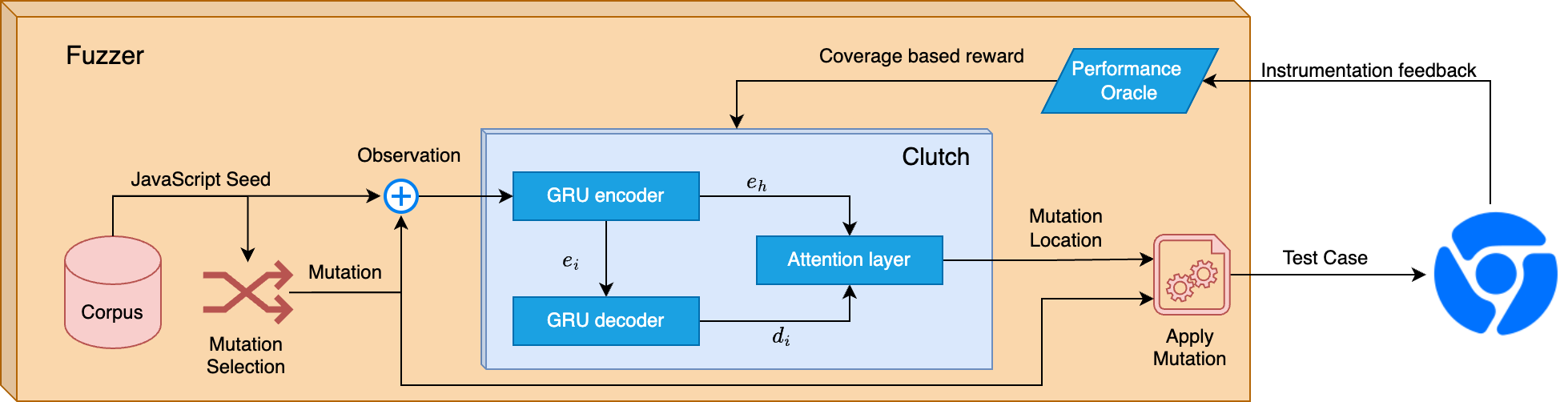}
    \caption{The \clutch\ architecture shown within a fuzzer.}
    \label{fig:clutch}
\end{figure*}

%
A high level system diagram of \clutch\ can be seen in Figure~\ref{fig:clutch}.
%
At each timestep, a test case is selected to mutate, which is represented for \clutch\ as a series of locations (observable arms). 
\clutch\ then observes the volatile number of arms and selects from the locations where to perform a mutation. 
%
The fuzzer then performs the mutation(s) at the selected location and submits this to the \js\ engine. 
Information on the performance of the test case is collected from the engine as it runs, which is used to compute the reward.
%
This is summarised in Algorithm~\ref{alg:clutch}.
%

\subsubsection{Model Architecture}
To handle the representation of \js\ that forms arms and contexts for \clutch\ we use a deep learning architecture. 
Specifically we use a sequence-to-sequence model with an attention mechanism in the form of a Pointer Network~\cite{vinyals_pointer_2017}. 
%
%
Using a Gated Recurrent Unit (GRU) model as an encoder-decoder allows the encoder and decoder states $e_1,\dots,e_n$ and $d_1,\dots,d_m$, to be used to compute an \emph{attention} vector $a^t$ over the locations available at each timestep ($t$), where parameters $v^T$, $W_1$ and, $W_2$ are learnable.
%
\begin{equation}\label{eq:pointer_attention}
\begin{split}
& u_j^t = v^T tanh(W_1 e_j + W_2 d_j) \ \  \ \  j \in (1,\dots,n)  \\
& a^t = \mathrm{softmax}(u^t )
\end{split}
\end{equation}
%
%
%
As a result, \clutch\ can take as input the IR of a \js\ test case, that constitutes a variable number of locations, and output an attention value for each location (vector $a^t$). The location with the highest value (estimated reward) can then be selected as the location to perform the mutation.

\begin{algorithm}[tb]
   \caption{Clutch approach to fuzzing.}
   \label{alg:clutch}
\begin{algorithmic}
   \STATE {\bfseries Input:} test cases $J$, mutations $M$, number of mutations to apply $n$, learning rate $\alpha$, update step $T$
   \STATE Initialize bandit $clutch$ with random weights $\theta$
   \REPEAT
   \STATE Initialize memory $K$
   \FOR{$t=1,2,\dots, T$}
   \STATE With uniform probability select test case $j_t \in J$
   \STATE With uniform probability select mutation $m_t \in M$
   \STATE $a_t, \varphi_t, r_t = clutch(j_t, m_t)$
    
   \STATE $l_t=\text{argmax}_n(a_t)$
   \STATE Apply mutation $m_t$ to $j_t$ at $l_t$ creating $l^\prime_t$
   \STATE Run testcase $l^\prime_t$, observing reward $R_t$

   \STATE Store ($l_t$, $r_t$, $\varphi_t$, $R_t$) in $K$
  
   \ENDFOR
   \STATE Update $\theta$ to minimise the loss in Eq.~\ref{eq:loss} using gradient descent with $\alpha$. 
   
   \UNTIL{done}
\end{algorithmic}
\end{algorithm}

The last challenge is the rate of exploration that \clutch\ should use to learn over time. 
In deep Reinforcement Learning (RL) models this often handled by hyperparameter $\varepsilon$, which indicates the rate of selecting random actions.
Conversely, deep MABs sample over a Gaussian distribution based on either the network parameters, or estimated rewards~\cite{zhang_neural_2021,hwang_combinatorial_2023}.
%
However, hyperparameter search and sampling network parameters is computationally expensive.
%
While it is possible to sample over the estimated rewards directly, this may not be sufficiently optimistic, resulting in lack of exportation and poor performance.

Instead, we can place a prior over the network, that can be tuned using Concrete Dropout~\cite{gal_concrete_2017}. 
In doing so, dropout parameter $p$ can be treated as a parameter to be trained via gradient descent.
Thus, the exploration-exploitation trade-off can be controlled via re regularisation parameters $r^1,\dots,r^4$.
As training increases and more data is seen, the dropout regularisation loss reduces as the uncertainty on the weights reduces. This reduces the exploration, resulting in more exploitation. 
Equally, the network can have higher uncertainty in the weights which will increase the regularisation, effectively increasing the exploration.
For more detailed explanation of Concrete Dropout, see Appendix~\ref{app:concrete}.
Thus, using Eq.~\ref{eq:pointer_attention}, we can form the output variables of \clutch.
The encoder and decoder outputs are passed through the respective hidden layers $W_1$, $W_2$, with Concrete Dropout applied.
%
This produces the attention vector $a_t$, which represents the estimated rewards of each location, used to select the location as in Algorithm~\ref{alg:clutch}.
The model also estimates the log variance, $\varphi_t$, used in a heteroskedastic loss as in Eq.~\ref{eq:loss}.
\begin{equation}\label{eq:clutch}
\begin{split}
a_t = &\  v^T\otimes (e_t\otimes W_1+ d_t\otimes W_2) \\
\varphi_t = &\  v^m\otimes (e_t\otimes W_1+ d_t\otimes W_2) \\
\end{split}
\end{equation}
Where $\otimes$ represents multiplication and Concrete Dropout. We can then determine the loss function for \clutch\ to include the regularisation terms of each layer ($r^1\dots r^4$):
\begin{equation}\label{eq:loss}
    l = \frac{1}{T}\sum^T_{t=1} (e^{-\varphi_t} + (R_t-a_t)^2 + \varphi_t ) +  \sum^T_{t=1} r^1_t + r^2_t + r^3_t + r^4_t 
\end{equation}




%
%



\subsection{Arms and Contexts}
%
\clutch\ will select arms that correspond to the location we can mutate within the test case. 
%
%
As we have outlined it is important for \clutch\ to understand specifics of the \js\ that will be mutated and the mutation that will be applied. 
Thus, at each timestep we give \clutch\ an observation of the available locations that can be selected and the mutation that will be performed. 
%

Arms and their contexts are, for \clutch,  agnostic of the fuzzer for which it will select locations.
%
This has the added advantage of reducing the overhead when selecting locations without further processing.
For instance, the IR used by \fz\ provides instructions that correspond to locations that can be selected.
For \clutch, each instruction can be uniquely mapped into a vocabulary. 
The context of each arm is then the vocabulary representation of the IR (the set of locations), in addition to the mutation that will be applied.

\subsection{Reward} 

Prior work from Li \etal~\cite{li_alphaprog_2022} studied the use of several key metrics for rewards when fuzzing a simple compiler, including: coverage, syntactic validity, and complexity of test cases.
Later work from Bates \etal~\cite{bates_reward_2023} and Foley and Maffeis~\cite{foley_apirl_2025} have
discussed the difficulty in designing rewards for real-world tasks, arguing for rewards consistent across training to ensure that diminishing returns do not effect the learnt policy. 
%
This is particularly relevant when using coverage based rewards where a) the total possible coverage can be quite large, and b) the majority of coverage is often found early in testing. 
Thus,  we design a reward, $R$ for \clutch\ that is consistent across training:
\begin{equation}\label{eq:reward}
    R =
    \begin{cases}
     1 + \frac{CC_{t}}{CC_{max}} + |\max(\mathrm{Var}(B_t) - \mathrm{Var}(B_{t-1}))| , & \parbox[t]{.3\textwidth}{Syntactically Correct Test} \\

     
    

     -1, & \parbox[t]{.1\textwidth}{Otherwise} \\
    \end{cases}
\end{equation}
%


Where $CC$ is the cyclomatic complexity (CC) and $B_t$ represents the vector of branch coverage at time $t$. 
This reward function balances validity, complexity, and coverage.
When a test case is not syntactically valid it penalises the choice, with a negative reward of $-1$. 
When test cases are valid, it rewards $1$ for the validity, additionally rewarding for how complex the test case is, as a ratio of current CC and the maximum CC seen historically. 

Finally, Eq~\ref{eq:reward} rewards for coverage using a term that incentivises the agent to explore new branches.
At each timestep we compute an updated running variance of each branch ($b\in B$) representing this as a vector: $\mathrm{Var}^{t}_{i=0}(B_i)$~\cite{caussinus_updating_1982}.
%
In doing so we can estimate the variance of each branch without retaining all the historic branch coverage. 
We then take the difference in variance between time $t$ and $t-1$, isolating the difference in variance (novelty in coverage) from the test case at time $t$.
The maximum value indicates the largest difference in coverage compared to the previous timestep, \ie\ where branches which are less frequently covered.
We then take the absolute value to reward for changing the variance.
Meaning that \clutch\ is incentivised not to hit the same branches frequently, but instead to explore other branches, including lower frequency branches.
Compared with traditional CBs where a reward is given per arm, we reward for the set of arms, as the reward is based on the overall test case performance, not individual arms.




\section{Implementation Details.}
The implementation of this architecture is as follows. 
Program representations are passed through a single embedding layer.
This embedding is then passed through a single GRU layer of width 64, to produce both encoder states and the terminal hidden state.  
The terminal hidden state is used to initialise the decoder, and give the decoder output.
The encoder and decoder output are passed through the respective hidden layers $W_1$, $W_2$  (of size 512) with concrete dropout applied.
%

We implement \clutch\ in Python 3.11 using PyTorch. 
We use the PythonKit \cite{vieito_pvieitopythonkit_2024} framework to interface \clutch\ with the fuzzers above (implemented in Swift), so they can share \js\ representations, coverage information, and locations.
Locations in each of these models are represented by the IR instructions found in each of the three fuzzers we place \clutch\ into. 
We run experiments on two identical machines running Ubuntu 24.04 with Intel i7 8700k and 64GB RAM.

\section{Field Test Evaluation}\label{eval:ftc}

In this Section we evaluate the effectiveness of \clutch\ at improving the efficiency of state-of-the-art fuzzers. Finally, we perform an bandit comparison study of \clutch\ in different settings. 
Interested readers may see additional analysis in Appendix~\ref{app:further exp}.

\begin{table*}
\centering
\caption{Correctness of generated test case and Branch coverage per generated test case (five run average) from 24h fuzzing of \clutch\ and baseline fuzzers.}
\label{tab:correct}
\resizebox{\textwidth}{!}{
\begin{tabular}{@{}cr||cc|cc|cc||cc|cc|cc@{}}
\toprule
\multicolumn{1}{l}{\multirow{2}{*}{Engine}} & \multicolumn{1}{l||}{\multirow{2}{*}{Metric}} & \multicolumn{6}{c||}{Correctness} & \multicolumn{6}{c}{Branch Coverage per test case} \\
\multicolumn{1}{l}{} & \multicolumn{1}{l||}{} & \fj & \fjc & \fz & \fzc & \jp & \jpc & \fj & \fjc & \fz & \fzc & \jp & \jpc \\ \midrule

\multirow{4}{*}{V8} & Average & 66.4\% & \textbf{80.5\%} & 65.9\% & \textbf{79.3\%} & 73.3\% & \textbf{85.8\%} & 0.566 & \textbf{0.627} & 0.382 & \textbf{0.387} & 0.514 & \textbf{0.585} \\

  &  Relative Change & \multicolumn{2}{c|}{8.321\% }& \multicolumn{2}{c|}{20.32\%} & \multicolumn{2}{c||}{8.585\%} & \multicolumn{2}{c|}{10.81\%} & \multicolumn{2}{c|}{1.152\%} & \multicolumn{2}{c}{13.72\%} \\
 & $\hat{A}_{12}$ & \multicolumn{2}{c|}{1.000} & \multicolumn{2}{c|}{1.000} & \multicolumn{2}{c||}{1.000} & \multicolumn{2}{c|}{0.560} & \multicolumn{2}{c|}{0.680} & \multicolumn{2}{c}{0.880} \\
 & $p$ & \multicolumn{2}{c|}{0.011} & \multicolumn{2}{c|}{0.002} & \multicolumn{2}{c||}{0.011} & \multicolumn{2}{c|}{0.099} & \multicolumn{2}{c|}{0.513} & \multicolumn{2}{c}{0.594}  \\ \midrule

\multirow{4}{*}{SpiderMonkey} & Average & 59.8\% & \textbf{77.4\%} & 64.9\% & \textbf{77.5\%} & 76.5\% & \textbf{89.5\%} & 0.050 & \textbf{0.067} & 0.059 & \textbf{0.082} & \textbf{0.092} & 0.081 \\

 &  Relative Change & \multicolumn{2}{c|}{ 9.843\% }& \multicolumn{2}{c|}{19.35\%} &\multicolumn{2}{c||}{8.615\%} & \multicolumn{2}{c|}{32.55\%} & \multicolumn{2}{c|}{38.56\%} & \multicolumn{2}{c}{-11.9\%} \\
 & $\hat{A}_{12}$ & \multicolumn{2}{c|}{1.0} & \multicolumn{2}{c|}{1.000} & \multicolumn{2}{c||}{1.000} & \multicolumn{2}{c|}{1.000} & \multicolumn{2}{c|}{1.000} & \multicolumn{2}{c}{0.222} \\
 & $p$ & \multicolumn{2}{c|}{0.008} & \multicolumn{2}{c|}{0.003} & \multicolumn{2}{c||}{0.036} & \multicolumn{2}{c|}{0.001} & \multicolumn{2}{c|}{0.001} & \multicolumn{2}{c}{0.287} \\\midrule
\multirow{4}{*}{JavaScriptCore} & Average & 70.5\% & \textbf{82.1\%} & 63.7\% & \textbf{78.7\%} & 71.8\% & \textbf{85.1\%} & \textbf{0.530} & 0.508 & \textbf{0.250} & 0.238 & 0.344 & \textbf{0.358} \\ 

 &  Relative Change & \multicolumn{2}{c|}{5.054\%} &\multicolumn{2}{c|}{23.52\%} & \multicolumn{2}{c||}{12.23\%} & \multicolumn{2}{c|}{-4.25\%} & \multicolumn{2}{c|}{-5.01\%} & \multicolumn{2}{c}{4.12\%} 
\\
 & $\hat{A}_{12}$ & \multicolumn{2}{c|}{1.000} & \multicolumn{2}{c|}{1.000} & \multicolumn{2}{c||}{1.000} & \multicolumn{2}{c|}{0.000} & \multicolumn{2}{c|}{0.200} & \multicolumn{2}{c}{1.000} \\
 & $p$ & \multicolumn{2}{c|}{0.008} & \multicolumn{2}{c|}{0.001} & \multicolumn{2}{c||}{0.008} & \multicolumn{2}{c|}{0.768} & \multicolumn{2}{c|}{0.055} & \multicolumn{2}{c}{0.096} \\ \bottomrule
\end{tabular}}
\end{table*}

\subsection{Experimental Setup}

\paragraph{Baselines.}
At its core, \clutch\ is a contextual bandit applied to select locations in variable length (volatile) JavaScript test cases to fuzz JavaScript engines.
%
We evaluate its effectiveness by inserting \clutch\ into three different state-of-the-art \js\ engine fuzzers: \fz~\cite{gros_fuzzilli_2023}, \fj~\cite{wang_fuzzjit_nodate}, and \jp~\cite{bernhard_jit-picking_2022}. We refer to these variants as \fzc, \fjc, and \jpc\ respectively. 
Each fuzzer applies the default number of $7$ mutations simultaneously, thus \clutch\ chooses $7$ arms at each timestep. 
%
%
We did not include OptFuzz~\cite{wang_optfuzz_2024}, another extension of \fz, in the comparison as the available prototype relied on a version of \code{llvm} incompatible with current operating systems and \js\ engines.
%

\paragraph{\js\ Engines.} 
We test \clutch\ on JavaScriptCore (used in Webkit and Safari), V8 (used in Chrome and Chromium), and SpiderMonkey (used in Firefox), the main \js\ engines in use, targeted by previous work from academics, security researchers, and project maintainers. 

%



\paragraph{Metrics.}
To assess the performance of \clutch\ we compare with the particular view to \emph{efficiency}.
%
%
We consider the \emph{correctness} of the test cases, \ie\ the test cases that are valid \js, not causing one of the four error types of \js: \code{SyntaxError},  \code{ReferenceError}, \code{TypeError}, and \code{RangeError}.
Generating \js\ that does not cause such errors results in test cases that pass initial semantic and syntactic checks; in turn facilitating the deeper testing into the less frequently used paths in the engine. 
%
A common metric to test the ability of a fuzzer to reach these paths is code \emph{coverage}.
%
To measure the \emph{efficiency} of \clutch\ we use the branch coverage per test case. 
Branch coverage is often used in the literature, providing a point of comparison with prior work~\cite{gros_fuzzilli_2023}.

\paragraph{Experiment Design.}
Each trial (\js\ engine and fuzzer) is run for 24 hours and repeated 5 times, overall constituting 3,240 CPU hours across all experiments. 
Repeating experiments reduces the effect of randomness and allows for the statistical testing of different fuzzing approaches, as advocated for by Schloegel \etal~\cite{schloegel_sok_2024}.
In particular, we use the Mann-Whitney U test to determine statistical significance, and Vargha and Delaney’s $\hat{A}_{12}$ test to quantify the magnitude of difference or effect size. 
The performance is considered significant when $p < 0.1$, and outperforms with a large effect size when $\hat{A}_{12} \geq 0.71$.






\subsection{Correctness.}\label{eval:ftc:correctness}
We first measure the ability of \clutch\ to improve the rate of generation of valid \js\ code, that is code that does not cause one of the four \js\ error types.
We report the validity results in Table~\ref{tab:correct}. 

All \clutch\ variants 
achieve a higher proportion of valid test cases compared to their non-\clutch\ counterparts. 
Indeed \clutch\ can quickly learn from its experiences to find strategies that improve the rate of valid test cases. 
Impressively, this performance is consistent across different fuzzing approaches, from differential testing to JIT-specific mutators. 
\clutch\ improves correctness  by at least 16.42\% (compared with \fj), and on average by 20.3\%. 
In all cases we show that these results are statistically significant, and have a large effect size. 

\jp\ and \jpc\ achieve the highest rate of correctness of baseline fuzzers, and \clutch\ variants, respectively.
%
\jp\ focuses on instrumentation and probing the engine using a subset of mutators from prior work~\cite{gros_fuzzilli_2023}, which causes this improvement in performance compared to the other approaches.

\begin{table*}[]
\centering
\caption{Total regret of combinatorial bandits in different settings.}
\label{tab:bandit_comp}
\resizebox{\textwidth}{!}{
\begin{tabular}{@{}ccccccccccc@{}}
\toprule
\multirow{2}{*}{Setting} & \multirow{2}{*}{
 \begin{tabular}[c]{@{}c@{}}Selected\\Arms\end{tabular}} & \multirow{2}{*}{Volatile} & \multicolumn{8}{c}{Total Regret} \\
 &  &  & ACC-UCB & CC-UCB & CombLinTS & CombLinUCB & CN-UCB & CN-TS & CN-TS(M=1) & \clutch\ \\\midrule
\multirow{2}{*}{Gowalla Dataset} & 2 & \checkmark & \begin{tabular}[c]{@{}c@{}}4915.64 \\ ($\pm$57.00)\end{tabular} & \begin{tabular}[c]{@{}c@{}}12328.73 \\ ($\pm$20.31)\end{tabular} & - & - & - & - & - & \begin{tabular}[c]{@{}c@{}}\textbf{1077.68} \\ \textbf{($\pm$962.21)}\end{tabular} \\
 & 4 & \checkmark & \begin{tabular}[c]{@{}c@{}}985.70 \\ ($\pm$20.35)\end{tabular} & \begin{tabular}[c]{@{}c@{}}3688.81 \\ ($\pm$15.15)\end{tabular} & - & - & - & - & - & \begin{tabular}[c]{@{}c@{}}\textbf{125.42} \\ (\textbf{$\pm$105.05)}\end{tabular} \\ \midrule
$R_{1}(\mathbf{x}) = \mathbf{x}^{\top}\mathbf{a}$ & 4 &  & - & - & \begin{tabular}[c]{@{}c@{}}2865.87 \\ ($\pm$321.02)\end{tabular} & \begin{tabular}[c]{@{}c@{}}2667.51 \\ ($\pm$227.48)\end{tabular} & \begin{tabular}[c]{@{}c@{}}74.43 \\ ($\pm$3.95)\end{tabular} & \begin{tabular}[c]{@{}c@{}}67.87 \\ ($\pm$2.12)\end{tabular} & \begin{tabular}[c]{@{}c@{}}70.80 \\ ($\pm$3.97)\end{tabular} & \begin{tabular}[c]{@{}c@{}}\textbf{65.04} \\ \textbf{($\pm$2.51)}\end{tabular} \\
$R_{2}(\mathbf{x}) = (\mathbf{x}^{\top}\mathbf{a})^{2}$ & 4 &  & - & - & \begin{tabular}[c]{@{}c@{}}1140.77 \\ ($\pm$23.56)\end{tabular} & \begin{tabular}[c]{@{}c@{}}1155.06 \\ ($\pm$14.31)\end{tabular} & \begin{tabular}[c]{@{}c@{}}310.01 \\ ($\pm$83.50)\end{tabular} & \begin{tabular}[c]{@{}c@{}}253.60 \\ ($\pm$28.48)\end{tabular} & \begin{tabular}[c]{@{}c@{}}242.26 \\ ($\pm$14.97)\end{tabular} & \begin{tabular}[c]{@{}c@{}}\textbf{216.54} \\ \textbf{($\pm$4.37)}\end{tabular} \\
$R_{3}(\mathbf{x}) = \cos(\pi \mathbf{x}^{\top}\mathbf{a})$ & 4 &  & - & - & \begin{tabular}[c]{@{}c@{}}2501.58 \\ ($\pm$35.68)\end{tabular} & \begin{tabular}[c]{@{}c@{}}2495.12 \\ ($\pm$48.78)\end{tabular} & \begin{tabular}[c]{@{}c@{}}948.33 \\ ($\pm$96.76)\end{tabular} & \begin{tabular}[c]{@{}c@{}}997.02 \\ ($\pm$158.80)\end{tabular} & \begin{tabular}[c]{@{}c@{}}991.70 \\ ($\pm$83.72)\end{tabular} & \begin{tabular}[c]{@{}c@{}}\textbf{325.86} \\ \textbf{($\pm$40.11)}\end{tabular} \\ \bottomrule
\end{tabular}}
\end{table*}
\subsection{Test Case Efficiency.}\label{eval:ftc:efficiency}


Recent work from Caturano \etal~\cite{caturano_discovering_2021} has argued that `more intelligent' tools achieve the same (or better) results with a given number of attempts. As such, we report the efficiency as coverage per test case in Table~\ref{tab:correct}.
This allows us to determine how \emph{efficient} each approach is with its test case budget. 
We can see that \clutch\ is generally able to improve the number of branches covered per execution, by an average of 8.9\%.
Indeed, \clutch\ increases the branches covered per execution \emph{consistently}, doing so in six of the nine combinations of fuzzer and \js\ engine. 
Specifically, each \clutch\ variant improves over the baseline fuzzer in two of the three engines. 
%
When comparing per \js\ engine we see that  \clutch\ always increases efficiency when fuzzing V8. 
%
This is due to a combination of the size of the engine and the resulting throughput of the fuzzer approaches. 
As the size of V8 is larger compared to JavaScriptCore and SpiderMonkey, the search space for \clutch\ is larger, giving it an advantage to target it efficiently. 
The throughput on V8 is also similar between \clutch\ variants and baseline fuzzers in comparison to the other \js\ engines, which also accounts for efficiency improvements of \clutch, as baseline fuzzers achieve less coverage per test case.
%

Both \fjc\ and \fzc\ have marginally lower efficiency in JavaScriptCore (-4.63\% on average), and significantly greater efficiency in SpiderMonkey (35.56\% on average).
Yet, \jpc\ does not follow this same behaviour, achieving 11.9\% fewer branches per execution on SpiderMonkey, and 4.12\% more on JavaScriptCore. 
This arises from the different approaches in fuzzing taken by \jp\ compared to \fz\ and \fzc, which leads to the highest rate of valid \js\ generation from both \jp\ and \jpc. 
As a result, \jp\ has a higher base efficiency compared to other fuzzers.


We can see from Vargha and Delaney’s $\hat{A}_{12}$ values that the effect size is often large. 
Similarly, we can see that when \clutch\ outperforms a fuzzer 
the result is often statistically significant, as indicated by $p < 0.1$. \mfhide{\mf{keep/bin?}}








\section{\clutch\ Selection}\label{eval:clutch}

We have seen that \clutch\ improves the efficiency of fuzzing by increasing the correctness of test cases, their coverage, and the coverage per test case.
However, it is important to consider the underlying cause. 
%
To investigate this we have recorded the mutations performed by baseline fuzzers and \clutch\ variants on 10,000 test cases. 
We capture where mutations are performed and run the test cases on V8 to capture branch coverage and \js\ errors. 
%
%
%

\begin{figure}[t!]
\centering
\subfigure[\fz\ V8 Mutations]{
    \includegraphics[width=0.46\linewidth]{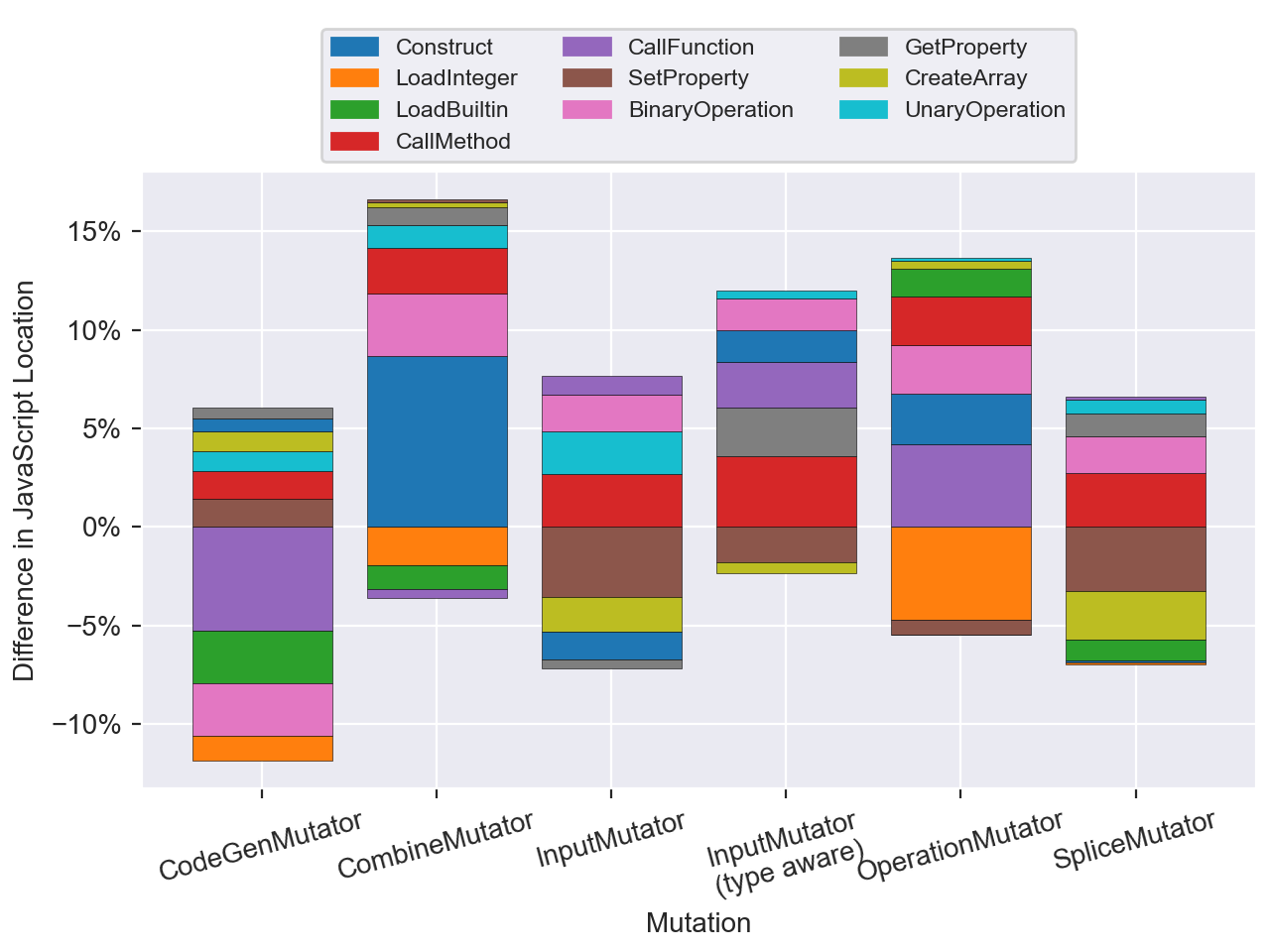}
    \label{fzc_v8_mut}}
\subfigure[\fz\ V8 Errors]{
    \includegraphics[width=0.46\linewidth]{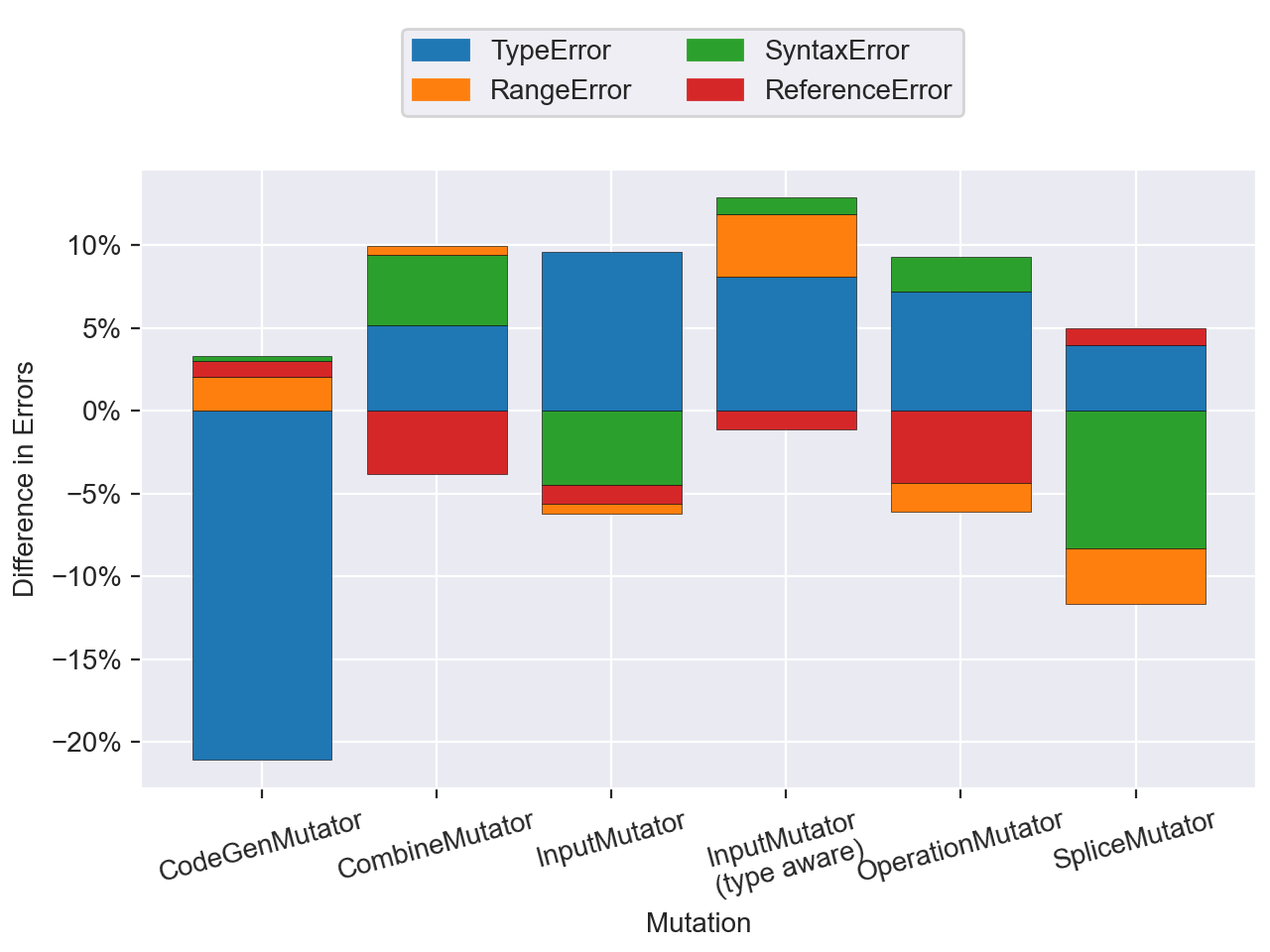}
    \label{fz_v8_err}}
\subfigure[\fj\ V8 Mutations]{
    \includegraphics[width=0.46\linewidth]{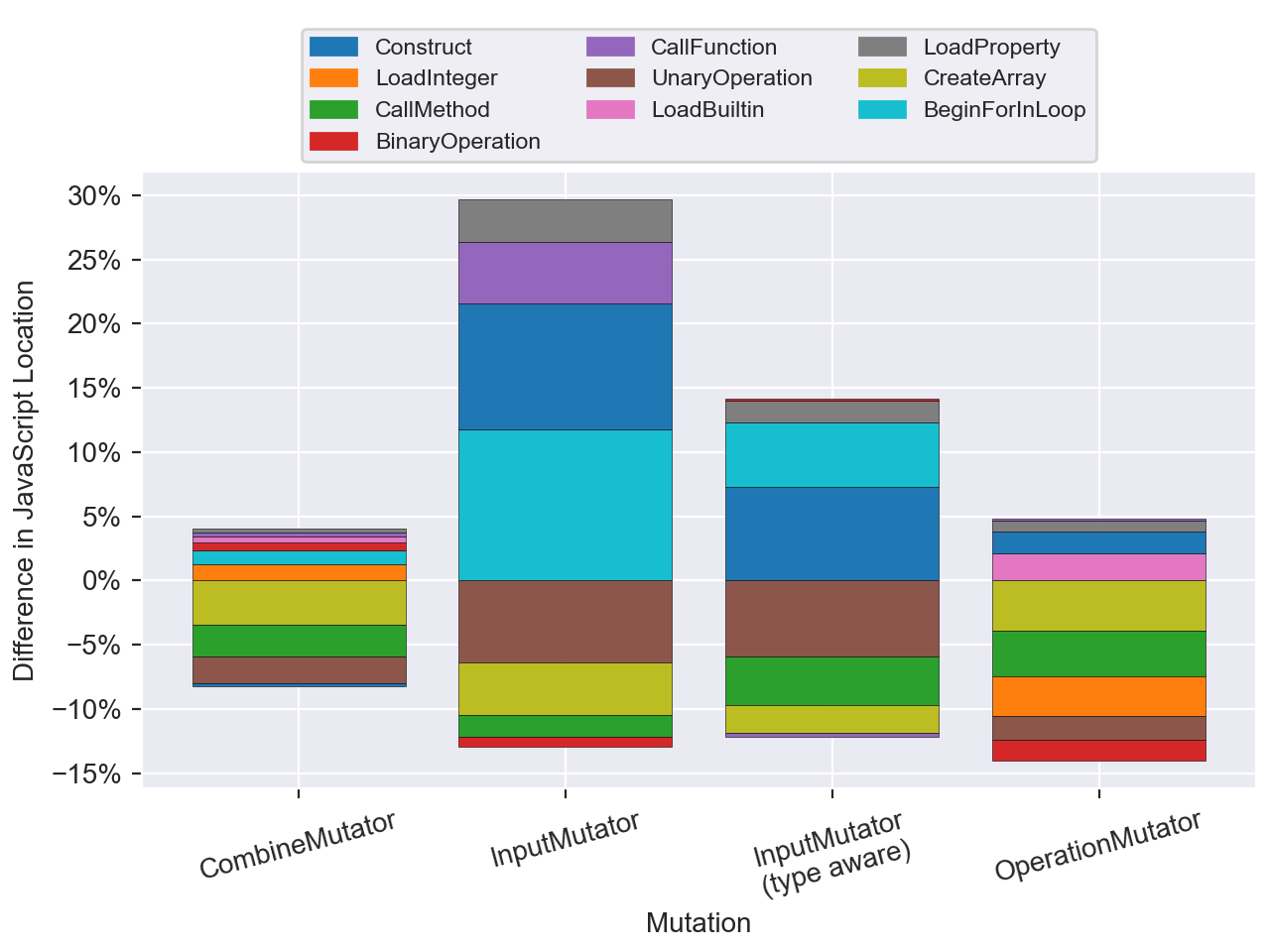}
    \label{fjc_v8_mut}}
\subfigure[\fj\ V8 Errors]{
    \includegraphics[width=0.46\linewidth]{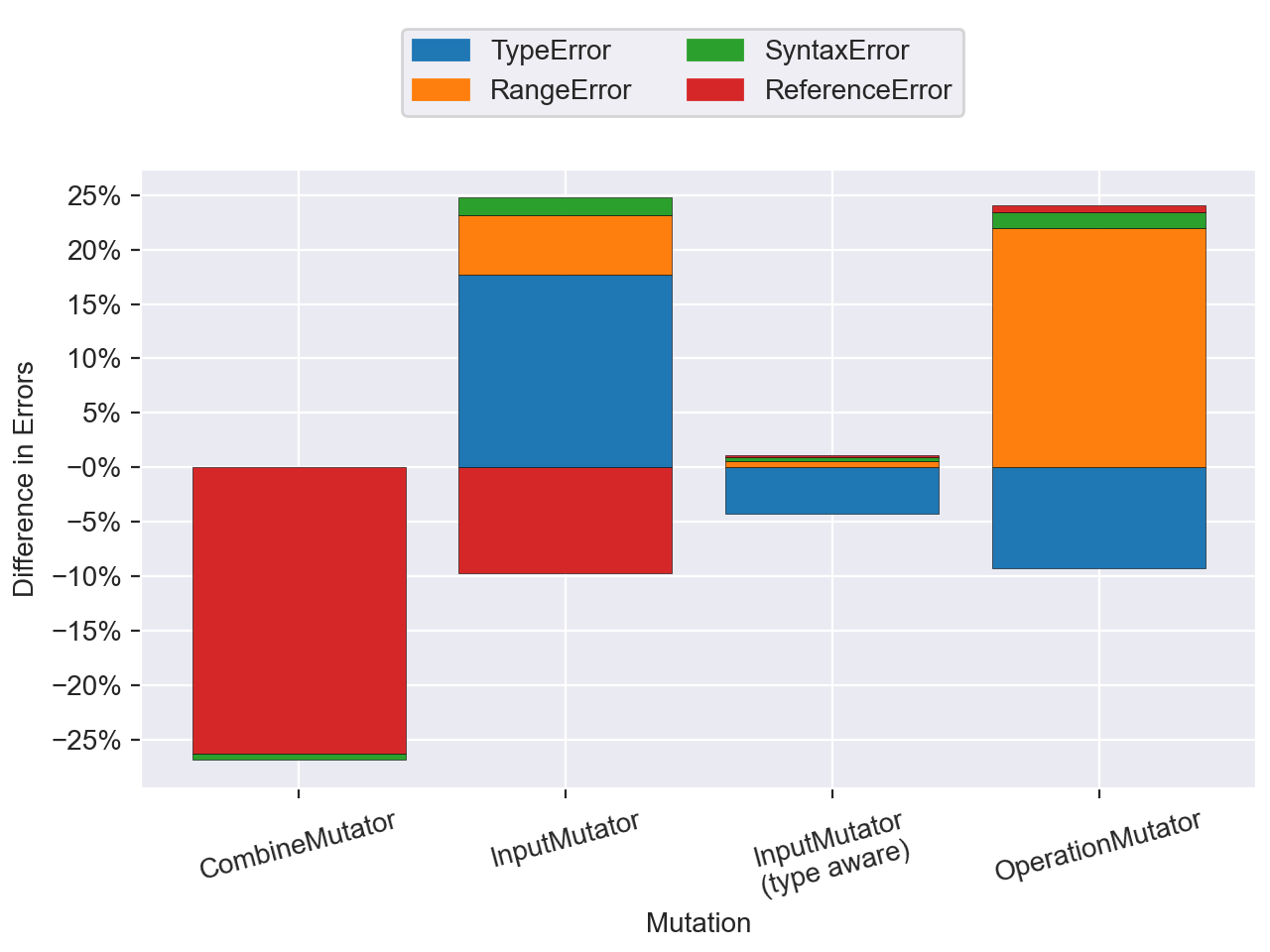}
    \label{fj_v8_err}}
\subfigure[\jp\ V8 Mutations]{
    \includegraphics[width=0.46\linewidth]{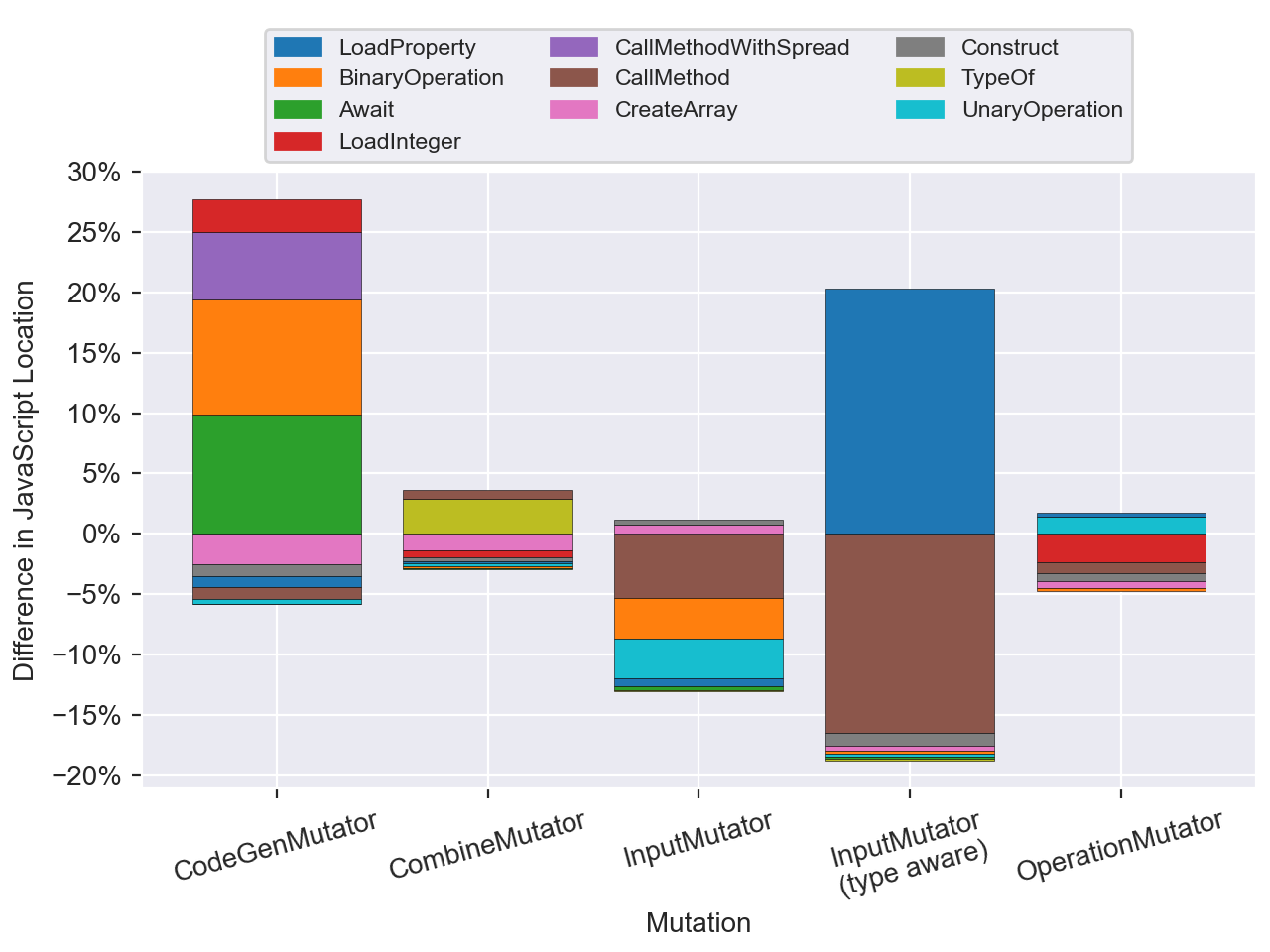}
    \label{jp_v8_mut}}
\subfigure[\jp\ V8 Errors]{
    \includegraphics[width=0.46\linewidth]{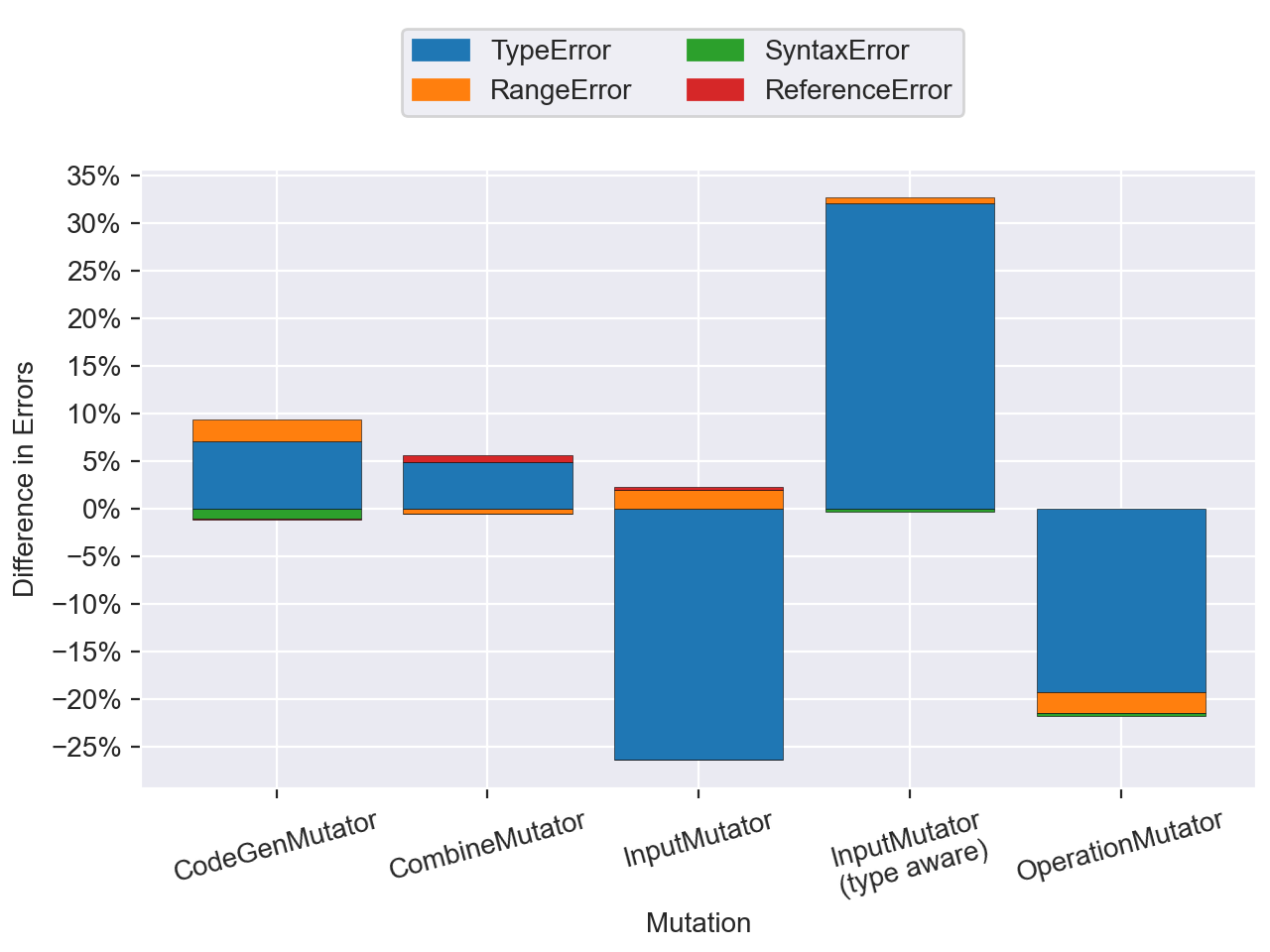}
    \label{jp_v8_err}}

\caption{Analysis of 10,000 mutated test cases by \clutch\ and the equivalent fuzzer when tested on V8. 
The left column shows the percentage difference from \clutch\ selection in its top ten selected instructions per mutation. 
%
The right column shows the difference between clutch and the baseline fuzzer in terms of the errors caused per mutation.}\label{fig:mutation_splits_v8}
\end{figure}

We provide the distribution of errors in Figure~\ref{fig:error_dist}.
Figures~\ref{fz_v8_err},~\ref{fj_v8_err},~\ref{jp_v8_err} show the difference in the errors across the mutations. 
Figures~\ref{fzc_v8_mut},~\ref{fjc_v8_mut}, and~\ref{jp_v8_mut} show
the distribution of IR instructions per mutation. 
%
%
For example Figure~\ref{fzc_v8_mut} shows the difference between \fz\ and \fzc. The \code{CodeGenMutator} bar shows where \clutch\ deviates from the strategy of \fz, \ie\ the \code{CallFunction} is selected 5.2\% less by \clutch, and \code{SetProperty} is chosen 1\% more. For readability we only report the top ten  IR instructions, thus the difference may not sum to zero for each bar.
%

\mfhide{\mf{this could go:}}
By comparing with the random selection of the fuzzers we can show what \clutch\ has learned to do. 
Importantly, we do so while accounting for the original distribution of the test cases, where some instructions, or mutations, may occur more frequently than others.
%
%
For instance, \code{InputMutator} cannot be applied to instructions with no input variables, such as a \code{BinaryOperation}.

\subsection{Distribution of Mutations}

We let \clutch\ observe the IR of a test case and the mutation that will be applied, with the intention for it to learn \emph{where} to apply different mutations. 
Signs of this occurring can be seen in Figure~\ref{fig:mutation_splits_v8} from the difference in the IR instructions chosen by \clutch\ for different mutations.
%
%
For instance, \code{BinaryOperation} is present in the top ten IR locations selected by \clutch\ regardless of the fuzzer approach. 
Yet, 
the preference of location to perform mutations varies across different \clutch\ variants. 
Consider the differences in selection between \jpc\ (Figure~\ref{jp_v8_mut}) and \fzc\ (Figure~\ref{fzc_v8_mut}). 
\jpc\ performs 9.8\% more \code{CodeGenMutator} mutations on \code{BinaryOperation}s than the random selection of \jp, where as \fzc\ has learnt an alternative strategy performing this combination 2.6\% less than \fz.
%
%

However, \clutch\ variants also perform mutations at similar locations, as six of the instructions are present in all variants. 
Consider \fjc\ and \fzc, which both favour the \code{Construct} instruction to perform mutations, indicated by its prevalence in Figure ~\ref{fzc_v8_mut} and~\ref{fjc_v8_mut}.
This seems a reasonable choice as \code{Construct} instructions create and initialise variables. 
%
By mutating these \clutch\ effects the structure and properties of test cases in an attempt to trigger additional functionality. 
%
%
%


%



\subsection{Distribution of Errors}

In Section~\ref{eval:ftc:correctness} we saw how \clutch\ increased the number of test cases that executed successfully. We investigate here why this is the case by studying what errors occur, and where. Figures~\ref{fz_v8_err},~\ref{fj_v8_err},~\ref{jp_v8_err} show the difference in the errors caused by test cases, and Figure~\ref{fig:error_dist} shows the distribution of errors across all fuzzers and \clutch\ variants.

Figure~\ref{fig:error_dist} shows a similar distribution of errors in the \clutch\ variants and baseline fuzzers, with some slight variation apparent between the different fuzzer types.
Overall, fuzzers cause \code{TypeError}s most frequently, which account for at least 78.2\% for \fzc, and in the worst case of \jp\ causing 92.1\% of errors.
This is not an unexpected distribution of errors as \js\ is a dynamically typed language where variable types can change type during the execution of a program. 
%
%
Combining this with the prevalence of mutations that change the type of a variable, either explicitly or implicitly increases the likelihood of a \code{TypeError}.

A more granular analysis of these errors, from Figures~\ref{fz_v8_err},~\ref{fj_v8_err},~\ref{jp_v8_err}, reveals a different story: errors occur because of different underlying causes.
%
These Figures show the difference in error types between \clutch\ variants and baseline fuzzers.
%
%
Consider Figure~\ref{fj_v8_err}, where \fjc\ causes fewer \code{ReferenceError}s when performing both \code{Input} and \code{Combine} mutations, leading to an overall reduction of these errors by 35\%.
%
%
However, while there is a commensurate increase in both \code{TypeError}s and \code{RangeError}s, the overall number of errors is reduced as seen in Section~\ref{eval:ftc:correctness}.
%
Such behaviour shows \clutch\ has learnt the syntax of test cases, selecting the most appropriate location to perform mutations. 
By comparison with Figure~\ref{fjc_v8_mut} we can see how applying mutations to different instructions causes a shift in errors. For instance, the use of \code{Construct} and \code{CallFunction} likely leads to an increase in \code{TypeError}s.

Similar comparisons can be made in the performance of \jpc.
In Figure~\ref{jp_v8_err}, the main difference in errors compared to \jp\ is from \code{TypeError}s.
We can see that the \code{InputMutator} (type aware) leads to an increase in \code{TypeError}s. 
When comparing with Figure~\ref{jp_v8_mut}, the cause of this is the shift in mutations; 
applying mutations more often to \code{LoadProperty}, and less to \code{CallMethod}.
The redistribution of mutations to different instructions also reduces the \code{TypeError}s in other locations: \code{InputMutator} and \code{OperationMutator}.
By doing this \clutch\ reduces the overall number of errors (Table~\ref{tab:correct}), and ratio of \code{TypeError}s (Figure~\ref{fig:error_dist}). 
%

\begin{figure*}
\subfigure[]{
    \includegraphics[width=0.5\linewidth]{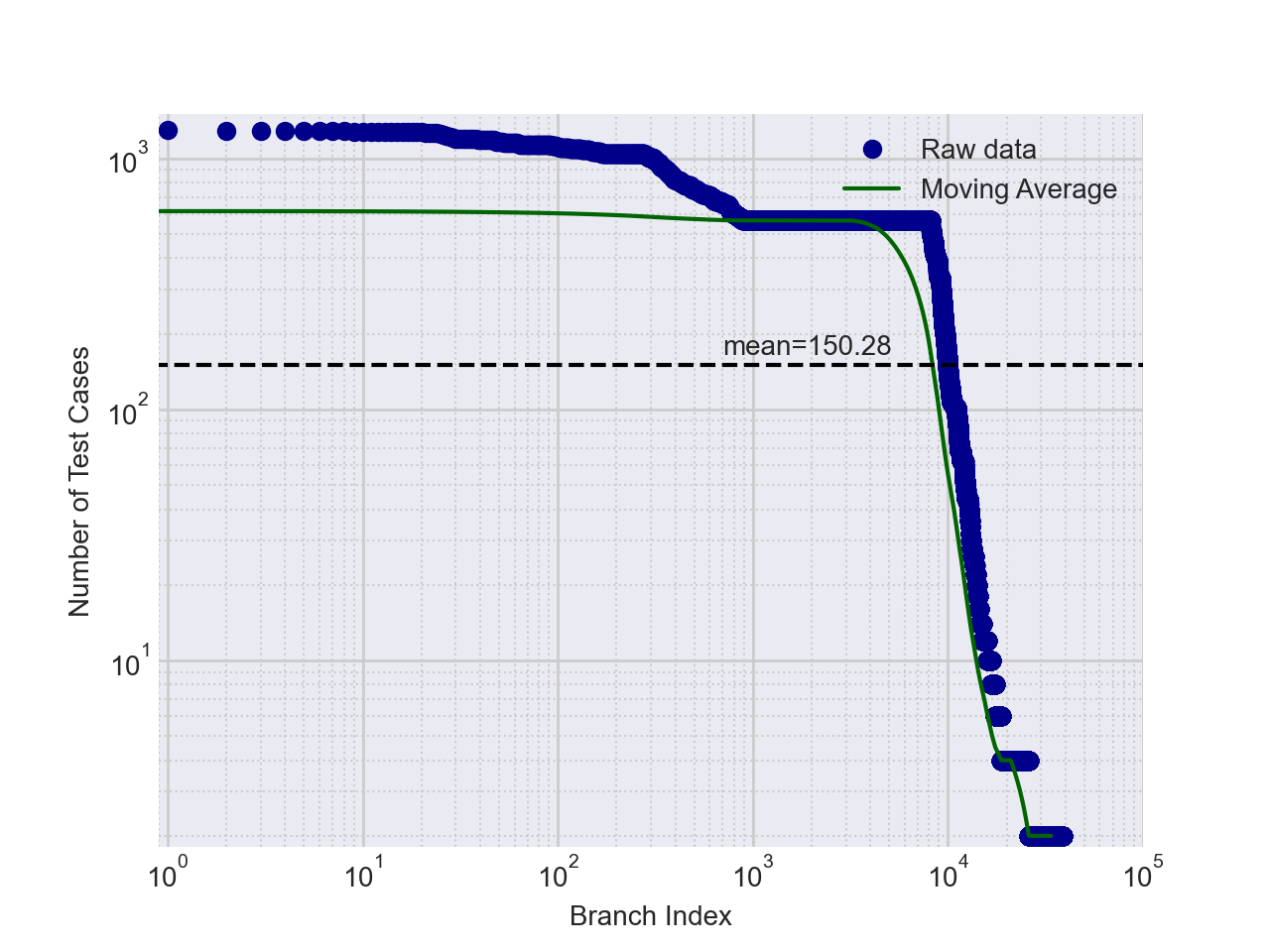}
    \label{cov_fz_v8}}
\subfigure[]{
    \includegraphics[width=0.5\linewidth]{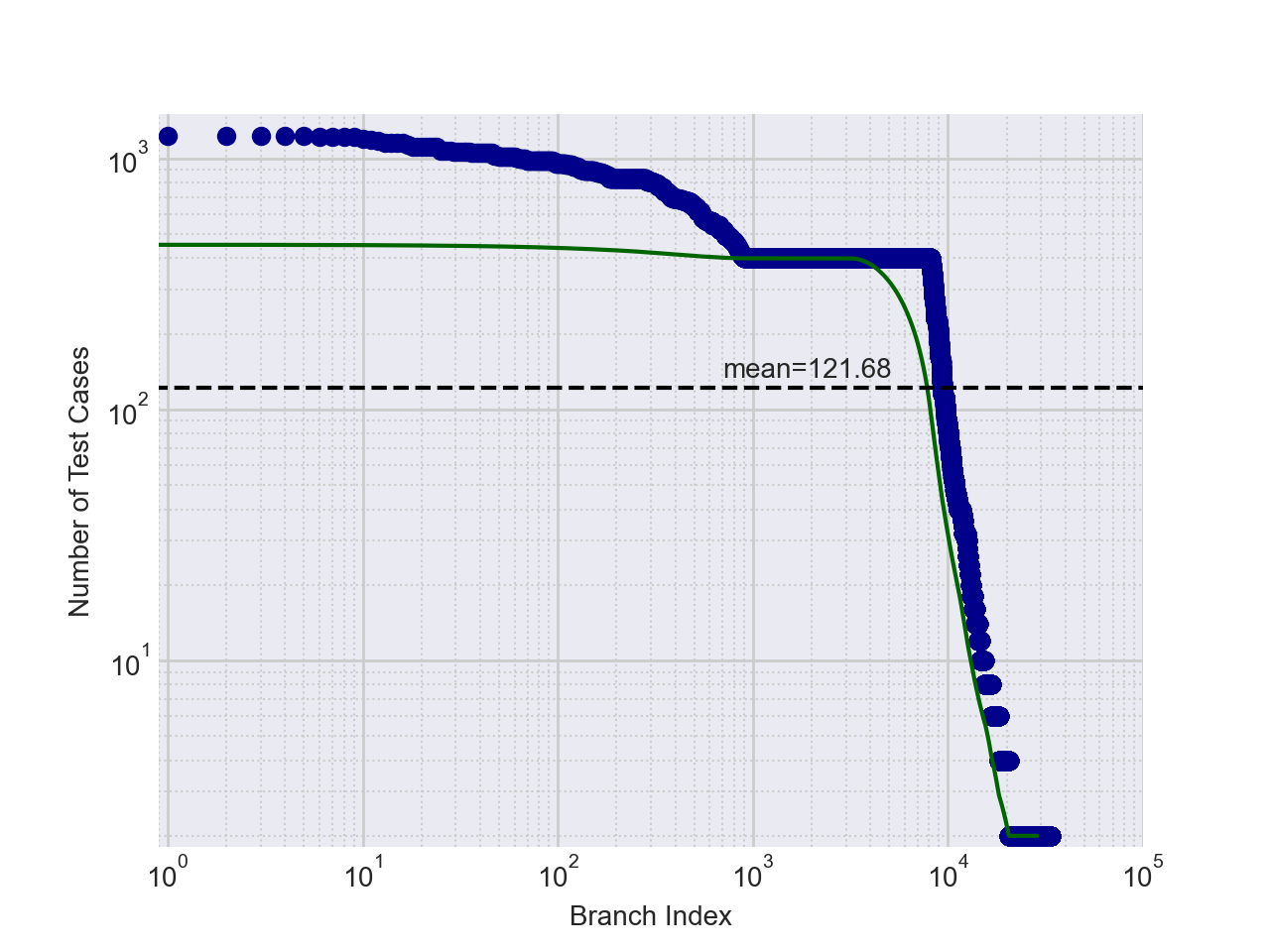}
    \label{cov_fzc_v8}}  
\caption{Frequency of branches in V8 that are reached from 10,000 test cases from \fz\ and \fzc.}\label{fig:coverage_splits}
\end{figure*}

\subsection{Frequency of Branch Coverage}

We have argued in favour of using \clutch\ to increase the efficiency of testing \js\ engines. 
%
%
To demonstrate this further we compare branch frequency from the 10,000 generated test cases generated by \fz\ and \fzc\ on V8. We report these results in Figure~\ref{fig:coverage_splits}. 
We can see that both \fz\ and \fzc\ find a comparable number of branches, covering 0.25\%. of the V8 engine, accounting for over 34,000 branches.
%
%
Comparable performance is also seen from the highest frequency of a single branch, 1320 for \fz\ and 1228 for \fzc. 
However, the frequency density of the the two highlights the difference in testing strategy.
On average \fz\ requires 150.28 test cases to cover each branch, 23.5\% more test cases than \clutch.
Test cases from \fz\ trigger the same shallow paths due to higher proportion of errors they contain. 
Indeed, such results show that \clutch\ explores a maximal number of low frequency paths~\cite{bohme_coverage-based_2016}.
%
\clutch\ does so by learning to perform mutations in locations which are, a) valid for \js, b) able to probe deeper in the engine more consistently.

\begin{figure*}[!ht]
\subfigure[]{
    \includegraphics[width=0.33\linewidth]{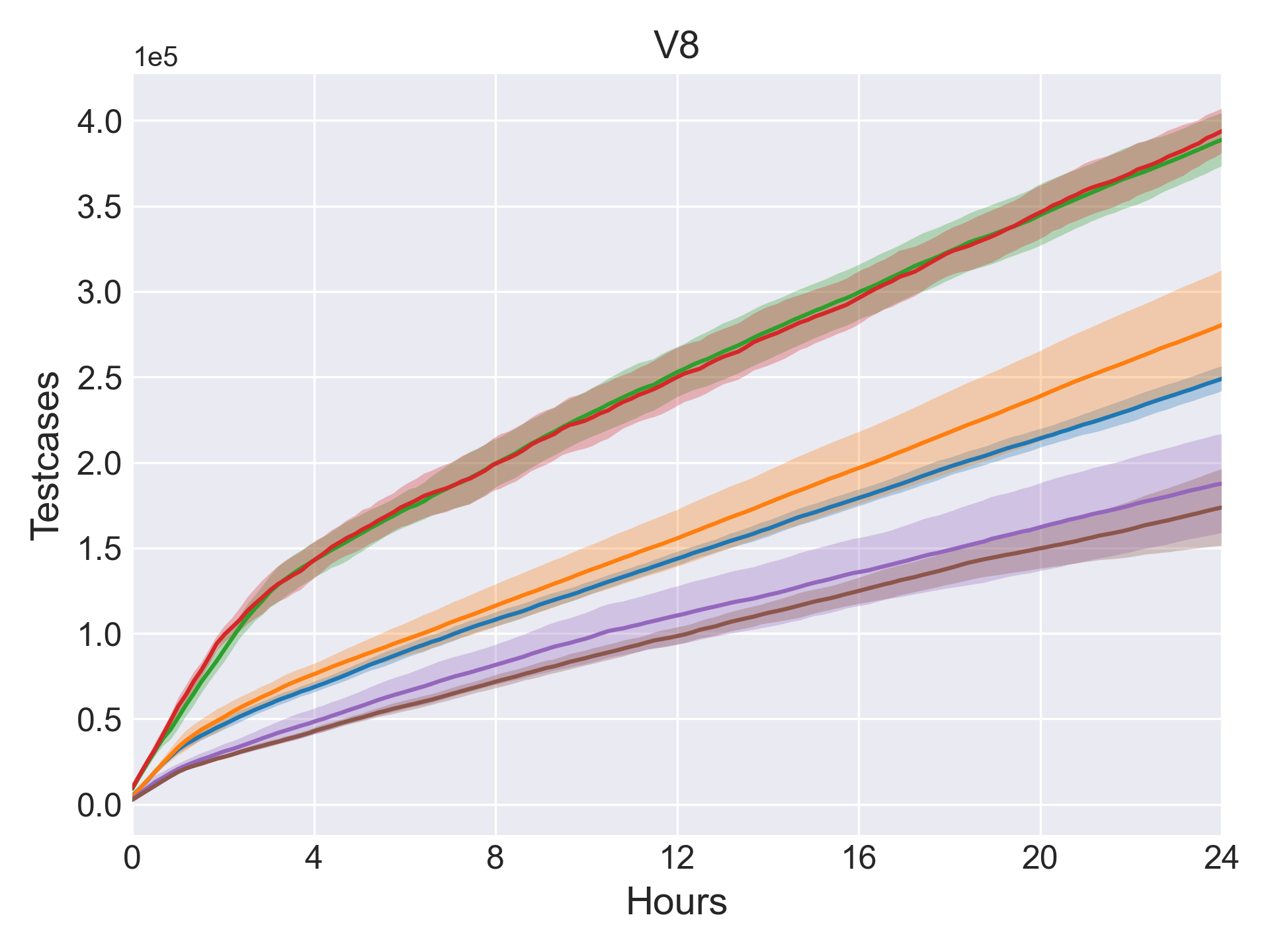}
    \label{v8_execs}}
\subfigure[]{
    \includegraphics[width=0.33\linewidth]{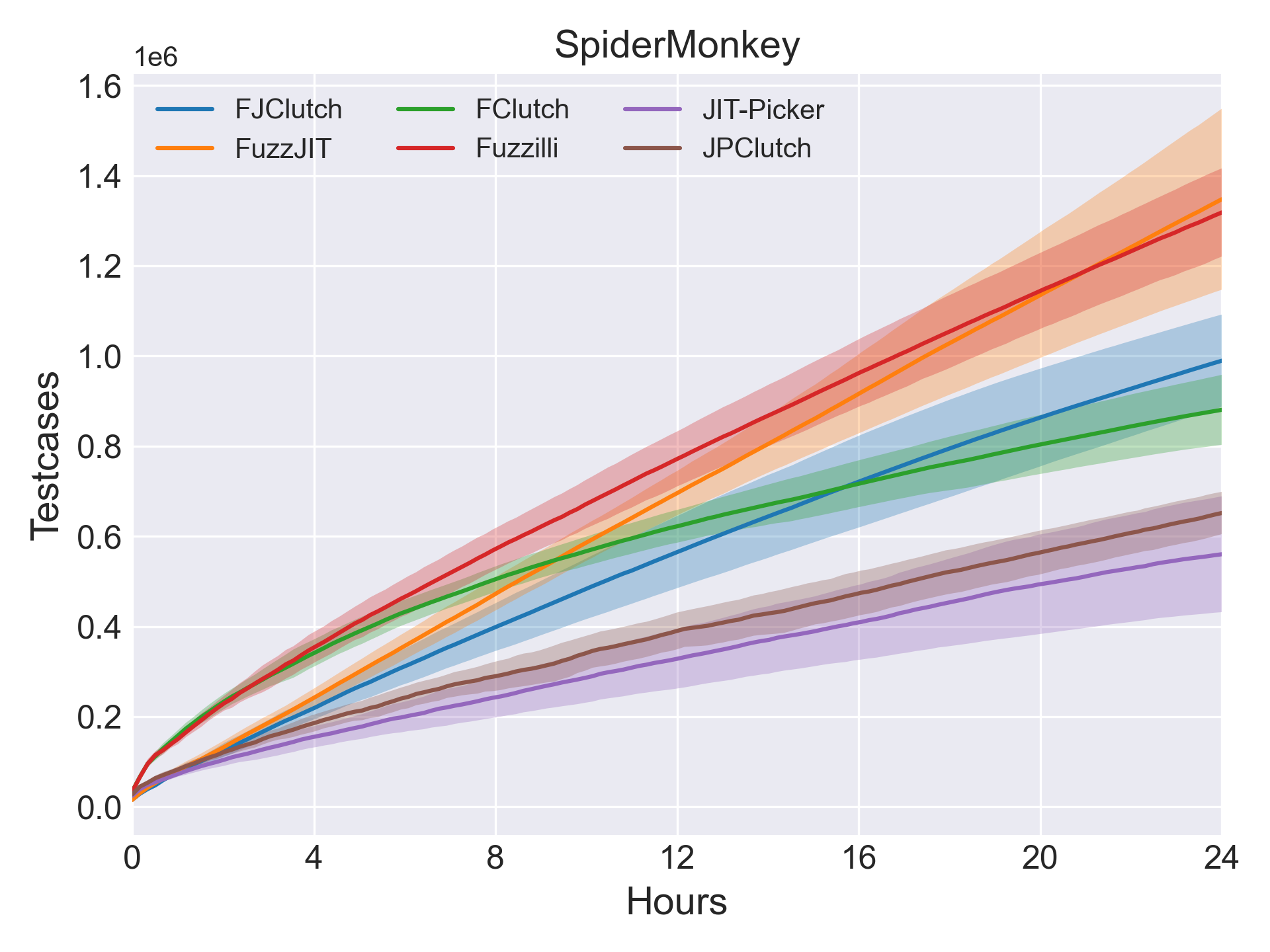}
    \label{sm_execs}}
\subfigure[]{
    \includegraphics[width=0.33\linewidth]{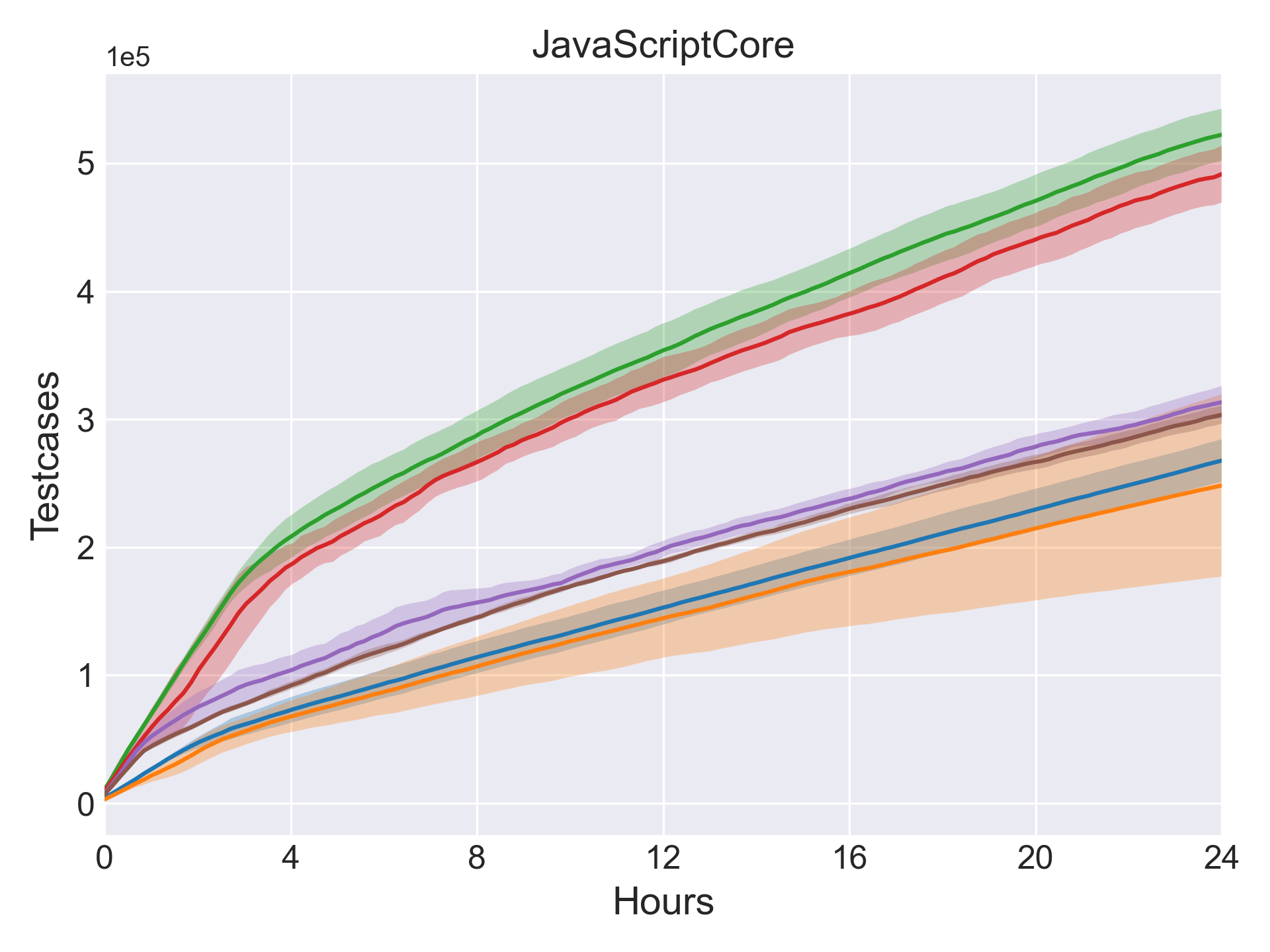}
    \label{jsc_execs}}
    
\caption{Throughput of test cases (five run average and standard deviation) from 24h fuzzing of \js\ engines.
}\label{fig:throughput}
\end{figure*}

\section{Combinatorial Bandit Evaluation}\label{sec:bandit_comp}


We must also consider the performance of \clutch\ in relation to alternative bandit approaches. 
However, to do so in the context of \js\ engines is challenging as it is computationally expensive to train bandits, and there is no ground truth to determine regret performance.
Instead, we use evaluation methodologies used in prior work, providing a ground truth reward, and relatively inexpensive training time.
Specifically, we consider the Gowalla Dataset in the same configuration as Nika \etal~\cite{nika_contextual_2020} for Volatile Combinatorial Bandits (VCBs), and use three hidden rewards as  Hwang \etal~\cite{hwang_combinatorial_2023} for CB.
%
%
We use the hyperparameters and experimental setup as in the prior work, running experiments for 10,000 rounds and repeating 10 times. We present the results in Table~\ref{tab:bandit_comp}. 


%
%

%
%
%


\paragraph{Volatile Arms.}
In this setting we consider a crowd sourcing problem used by Nika \etal~\cite{nika_contextual_2020}, using the Gowalla dataset. At each round $t$ a random number of user check-ins are sampled, where each check-in is an arm. Context is the location (normalised longitude and latitude lying in $[0,1]^2$), and the reward is the `willingness to work', \ie\ battery status. Bandits must then select $K\in{2,4}$ locations.
We compare against two state-of-the-art bandits that handle the volatile nature of the arms: ACC-UCB~\cite{nika_contextual_2020} and CC-UCB~\cite{chen_contextual_2018}. 
%
%
From Table~\ref{tab:bandit_comp} we see that bandits improve when selecting an increasing number of arms due increased information gathered from the additional arms.
Furthermore, \clutch\ has the best performance in both $K\in{2,4}$, achieving at least 78.1\% fewer regret than other bandits.
%
%


\paragraph{Non-Volatile Arms.}
In this setting we consider three different, unknown, reward functions: $R_{1}(\mathbf{x}) = \mathbf{x}^{\top}\mathbf{a}$, $R_{2}(\mathbf{x}) = (\mathbf{x}^{\top}\mathbf{a})^{2}$, and $R_{3}(\mathbf{x}) = \cos(\pi \mathbf{x}^{\top}\mathbf{a})$. Where $x$ is the set of contexts for arms and $a$ has the same dimension as $x$ and is randomly generated from a unit ball, remaining fixed during the experiment.
Bandits must select from $K=4$ arms at each round, observing a context of $80$ features. Such an evaluation is inline with Hwang \etal~\cite{hwang_combinatorial_2023} and Zhou \etal~\cite{zhou_neural_2020}.
We compare against contextual neural bandits CN-UCB and CN-TS, and their linear equivalents CombLinUCB and CombLinTS~\cite{hwang_combinatorial_2023}. 
We additionally include CN-TS(M=1), a special case of CN-TS drawing only one sample per arm. 
Volatile bandits require context features to be $N=2$, which not originally considered by Hwang \etal~\cite{hwang_combinatorial_2023} which uses 80 features ($N=80$).
For completeness we consider $N=2$ in Appendix~\ref{app:furtherbandit}.
%
From Table~\ref{tab:bandit_comp} we again see that the deep CBs outperform the linear counterparts. 
We also see that across all rewards \clutch\ outperforms the deep CBs, achieving between 4.1\%-67.3\% fewer regret.

Furthermore, across both volatile and non-volatile settings \clutch\ outperforms with large effect size ($\hat{A}_{12}\geq 0.78$) and statistical significance ($p<0.04$) (see Appendix~\ref{app:furtherbandit}). 


\section{Threats to validity}

The selection of hyperparameters in a learning architecture may depend on domain knowledge, leading to bias, and may in general affect the performance of the model unpredictably.
To mitigate this, \clutch\ controls its own rate of exportation, uses hyperparameters from prior work where applicable~\cite{gal_concrete_2017,collier_deep_2018,vinyals_pointer_2017}, and uses a grid-search to find optimal values for remaining hyperparameters (Appendix~\ref{app:hyperparam}). 

We did not place \clutch\ inside an exhaustive list of all \js\ engine fuzzers.
In particular, we have excluded OptFuzz~\cite{wang_optfuzz_2024} and Superion~\cite{wang_superion_2019}. 
OptFuzz, while publicly available, requires recompiling the \code{llvm} library with custom hooks, and does not fully document the dependencies required to do so.
Superion is used for fuzzing both XML and \js\ engines, extending the AFL fuzzing model. As such we considered the approach and underlying system out-of-scope.
Indeed, it is non-trivial to place \clutch\ inside new fuzzers, even those based on similar architecture.
Instead, we selected three state-of-the-art fuzzers to evaluate the performance of \clutch. 
We argue these fuzzers demonstrate the generalisability of our approach, as they use different approaches and testing mutations, ranging from targeting the JIT to differential testing.
We further test each fuzzer and \clutch\ variant on three separate and commonly used \js\ engines. This removes bias from test subjects, showing the real world application of \clutch.

To mitigate for randomness in the different approaches we conducted large scale experiments. We repeated experiments in Section~\ref{eval:ftc}, and Appendix~\ref{eval:accub} five times, accounting for 3,240 compute hours of experiments. 
In reporting our results we have followed guidance from
Schlosser \etal~\cite{schloegel_sok_2024}, reporting metrics which provide deeper statistical comparison between different sets of data. 
Additionally, in Section~\ref{eval:clutch}, we use 10,000 test cases to compare between \clutch\ and the fuzzer baselines to provide an in-depth study on their behavioural differences.

\section{Conclusion}

The use of fuzzing to find bugs and vulnerabilities in \js\ engines has become the default testing strategy. 
While existing research has focused on oracles, instrumentation, and heuristics for mutations, they perform mutations at random locations in test cases. 
We posit that a bandit model can \emph{learn} effective locations to perform mutations to increase testing efficiency. 
%
%
However, to do so must over come several challenges: 1) a bandit that can handle the high dimensional state-action space, 2) can adapt to the volatile number of arms that represent the variable locations in  \js\ test cases, 3) can dynamically adjust the rate of exploration during testing. 
%
Our approach, \clutch, overcomes these challenges with the attention mechanism found in pointer networks to hand the volatile number of arms, and concrete dropout to adjust the rate of exploration.
We place \clutch\ inside existing fuzzers, showing it can increase correctness of test cases and efficiency of coverage per test case. 
We further evaluate \clutch\ in volatile and combinatorial settings, showing state-of-the-art performance compared to existing approaches.
\clutch\ demonstrates impressive performance locating \emph{where} to perform mutations in a test case, highlighting how bandits can solve real-wold problems.


%
%

%
%
%
%
%
%
%
%
%

\newpage
\clearpage

\bibliography{refs}
\bibliographystyle{unsrt}

\newpage
\appendix
\onecolumn
%
%


\section{Hyperparameters}\label{app:hyperparam}

 Lower and upper bounds of the of the hyperparameters used in the grid search for \clutch\ are displayed in Table~\ref{table:hyperparam}. Values were sampled uniformly to determine the optimal hyperparameters. We also include the selected values. In Section~\ref{sec:bandit_comp} we compare against VCB and CB bandits using evaluation benchmarks taken from prior work; thus we use the same hyperparameters as in prior work. 
 
                \begin{table}[!ht]
        \centering
        \caption{Grid search ranges for hyperparameters neural network models\label{table:hyperparam}}
        \centering 
        {\footnotesize
\begin{tabular}{@{}lccc@{}}
\toprule
 Hyperparameter & Lower Bound & Upper Bound & Selected \\ \midrule
 $\gamma$ & 0.5 & 0.999 & 0.9 \\
  Learning Rate & 0.05 & 0.0005 & 0.005 \\
  Update Step & 10 & 10000 & 200 \\ 
  Update Type & Exponential & Linear & Linear \\ \midrule

\end{tabular}}
        
        \end{table}

\section{Concrete Dropout}\label{app:concrete}
Dropout randomly samples from a masked weight matrix in the neural network~\cite{collier_deep_2018}. 
This dropout probability and weight matrix can be approximated to the probabilistic deep Gaussian process~\cite{damianou_deep_2013}. 
By taking this approximation we can apply a Bayesian deep learning model to a neural network of $L$ layers and $\theta$ variations parameters.
This allows us to place a prior on the network weights and then learn the posterior $q_\theta(\omega)$ distribution over the weights ($\omega=\{W_l\}^L_{l=1}$)~\cite{gal_concrete_2017}. This can be formally written as the objective functions:

\begin{equation}
    \hat{\mathcal{L}}_{MC}(\theta) = - \frac{1}M{}\sum_{i\in S}\log p(\mathbf{y}_i|\mathcal{\mathbf{f}}^{\omega}(\mathcal{\mathbf{x}}_i))+\frac{1}{N}\mathrm{KL}(q_\theta(\omega)||p(\omega))
\end{equation}

where $S$ is a set of $M$ data points, $N$ is the number of data points, and $\theta$ is the optimisable parameters,  $\mathcal{\mathbf{f}}^{\omega}(\mathcal{\mathbf{x}}_i)$ is the model output given $\mathbf{x}_i$, and $p(\mathbf{y}_i|\mathcal{\mathbf{f}}^{\omega}(\mathcal{\mathbf{x}}_i))$ is the model's likelihood. 

The KL term $\mathrm{KL}(q_\theta(\omega)||p(\omega))$ can be considered the `regularisation' term to ensure that the posterior $q_\theta(\omega)$ and prior $p(\omega)$ distributions do not diverge too far. 
Gal \etal~\cite{gal_concrete_2017} show that the KL term can be approximated to depend only on the probability of turning off a Bernoulli random variable (node) with  with probability $p$.
From this we can derive Concrete Dropout~\cite{gal_concrete_2017}, where we continuously relax dropout’s discrete mask. 
%
%
%


\begin{figure}
    \centering
    \includegraphics[width=\linewidth]{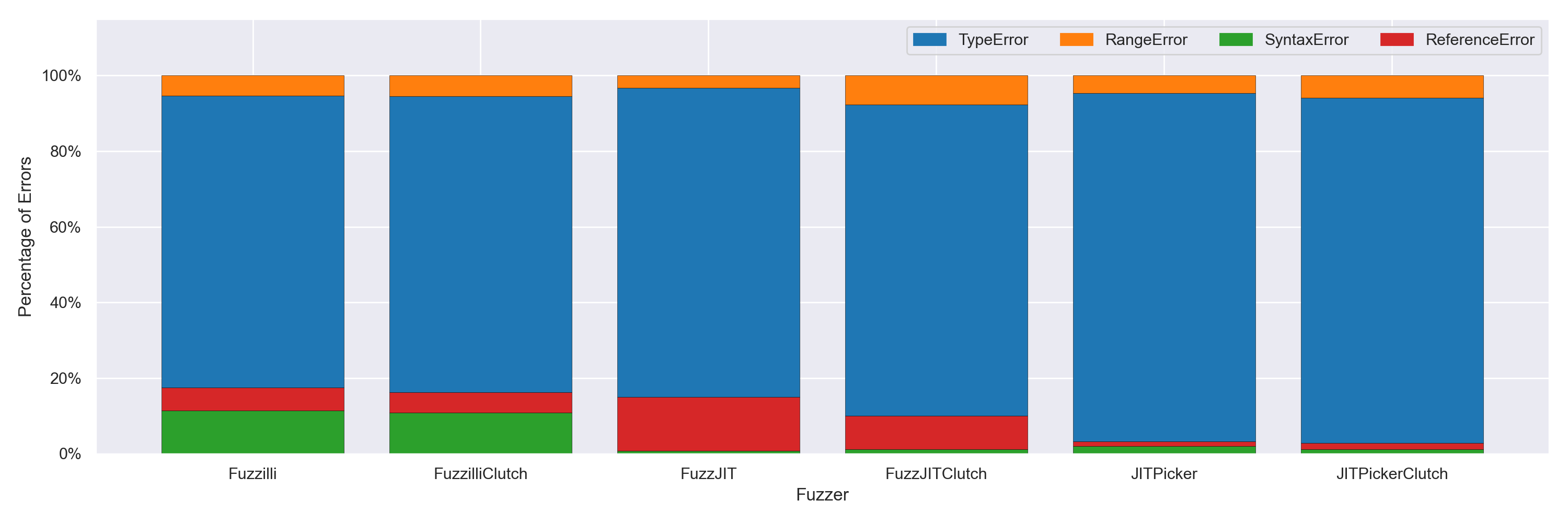}
    \caption{Distribution of error types caused by 10,000 test cases from different fuzzers, when testing on V8.}
    \label{fig:error_dist}
\end{figure}


\section{Additional Field Test Evaluation}\label{app:further exp}


\subsection{Throughput}
The throughput of a fuzzer is important to consider as, typically, the more test cases we execute the higher the likelihood of triggering a bug.  
We monitor the throughput of each fuzzer, reporting the results in Figure~\ref{fig:throughput}.
We would expect that \clutch\ variants will take longer to execute a single test case compared to their relative counterparts, as calling an external implementation of a neural network in Python will inevitably introduce a higher overhead compared to a random selection implemented directly in the fuzzer. 
%
%
Increasing the rate correctness of test cases may also reduce throughput: by having test cases that are valid they test deeper into the engine, and in doing so, they take longer to execute. 
%


Overall, Figure~\ref{fig:throughput} shows that each of the fuzzers have a different rate at which they execute test cases (similar to prior work), and that \clutch\ versions have a lower throughput.  

\begin{table*}
\centering
\caption{Branch coverage (five run average) from 24h fuzzing of \clutch\ and baseline fuzzers.}
\label{tab:coverage}
{\scriptsize
\begin{tabular}{@{}cr|cc|cc|cc@{}}
\toprule
Engine & Metric & \fj & \fjc & \fz & \fzc & \jp & \jpc \\ \midrule
\multirow{4}{*}{V8} & Average & \textbf{11.26\%} & 11.08\% & \textbf{10.69\%} & 10.67\% & 6.854\% & \textbf{7.212\%} \\
 & Relative Change & \multicolumn{2}{c|}{-1.62\%} & \multicolumn{2}{c|}{-0.16\%} & \multicolumn{2}{c}{5.224\%} \\
 & $\hat{A}_{12}$ & \multicolumn{2}{c|}{0.440} & \multicolumn{2}{c|}{0.640} & \multicolumn{2}{c}{0.640} \\
 & p & \multicolumn{2}{c|}{0.099} & \multicolumn{2}{c|}{1.000} & \multicolumn{2}{c}{0.953} \\ \midrule
\multirow{4}{*}{SpiderMonkey} & Average & \textbf{21.30\%} & 20.73\% & \textbf{24.50\%} & 22.70\% & 16.18\% & \textbf{16.97\%} \\
 & Relative Change & \multicolumn{2}{c|}{-2.68\%} & \multicolumn{2}{c|}{-7.33\%} & \multicolumn{2}{c}{4.898\%} \\
 & $\hat{A}_{12}$ & \multicolumn{2}{c|}{0.000} & \multicolumn{2}{c|}{0.000} & \multicolumn{2}{c}{0.667} \\
 & p & \multicolumn{2}{c|}{0.055} & \multicolumn{2}{c|}{0.001} & \multicolumn{2}{c}{0.112} \\ \midrule
\multirow{4}{*}{JavaScriptCore} & Average & 17.64\% & \textbf{18.21\%} & 16.49\% & \textbf{16.64\%} & \textbf{14.58\%} & 14.55\% \\
 & Relative Change & \multicolumn{2}{c|}{3.229\%} & \multicolumn{2}{c|}{0.923\%} & \multicolumn{2}{c}{-0.18\%} \\
 & $\hat{A}_{12}$ & \multicolumn{2}{c|}{1.000} & \multicolumn{2}{c|}{0.680} & \multicolumn{2}{c}{0.520} \\
 & p & \multicolumn{2}{c|}{0.859} & \multicolumn{2}{c|}{0.594} & \multicolumn{2}{c}{0.755} \\  \bottomrule
\end{tabular}}
\end{table*}

\subsection{Coverage}


\begin{figure}
\subfigure[]{
    \includegraphics[width=0.33\linewidth]{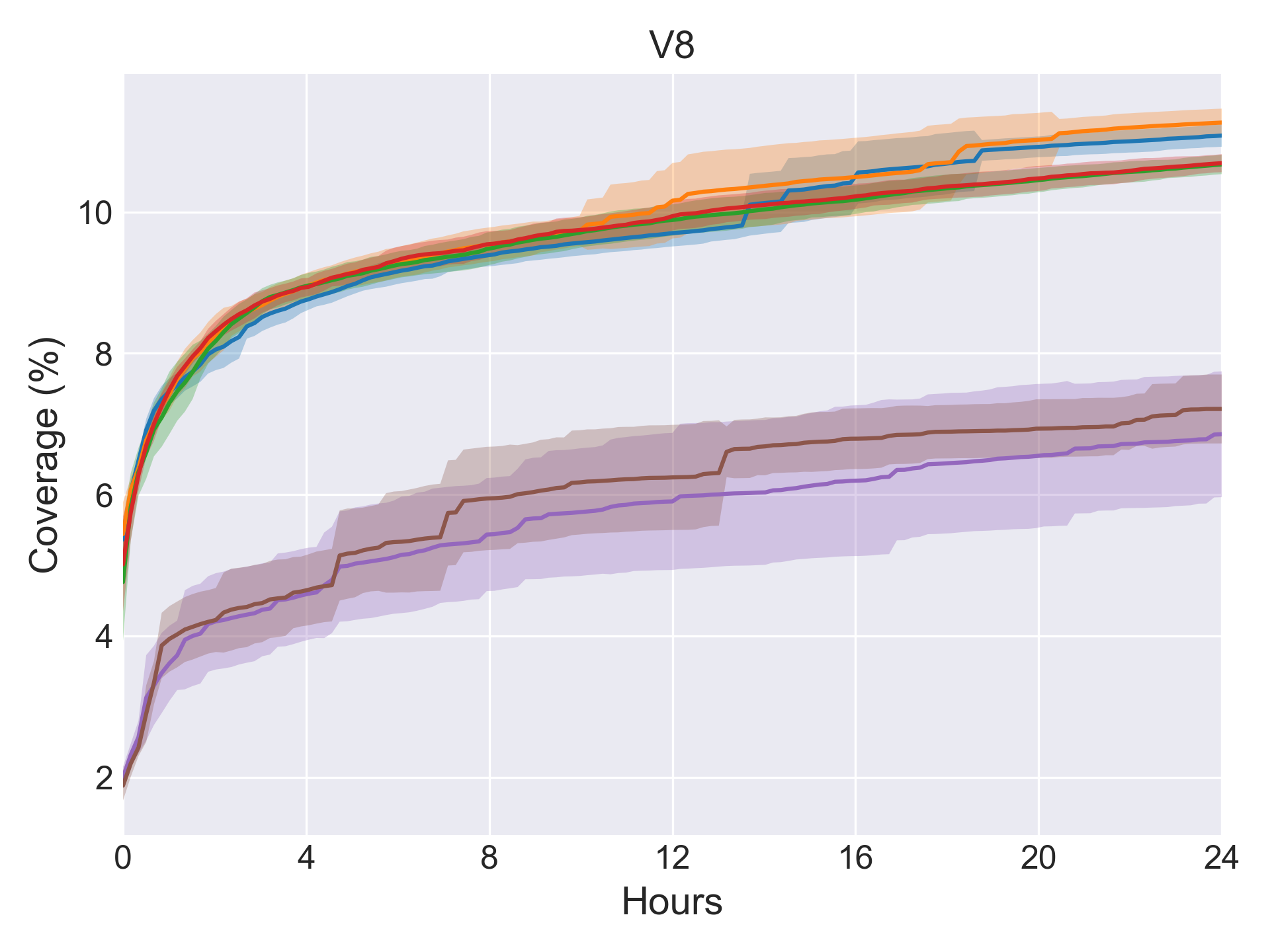}
    \label{v8_cov}}
\subfigure[]{
    \includegraphics[width=0.33\linewidth]{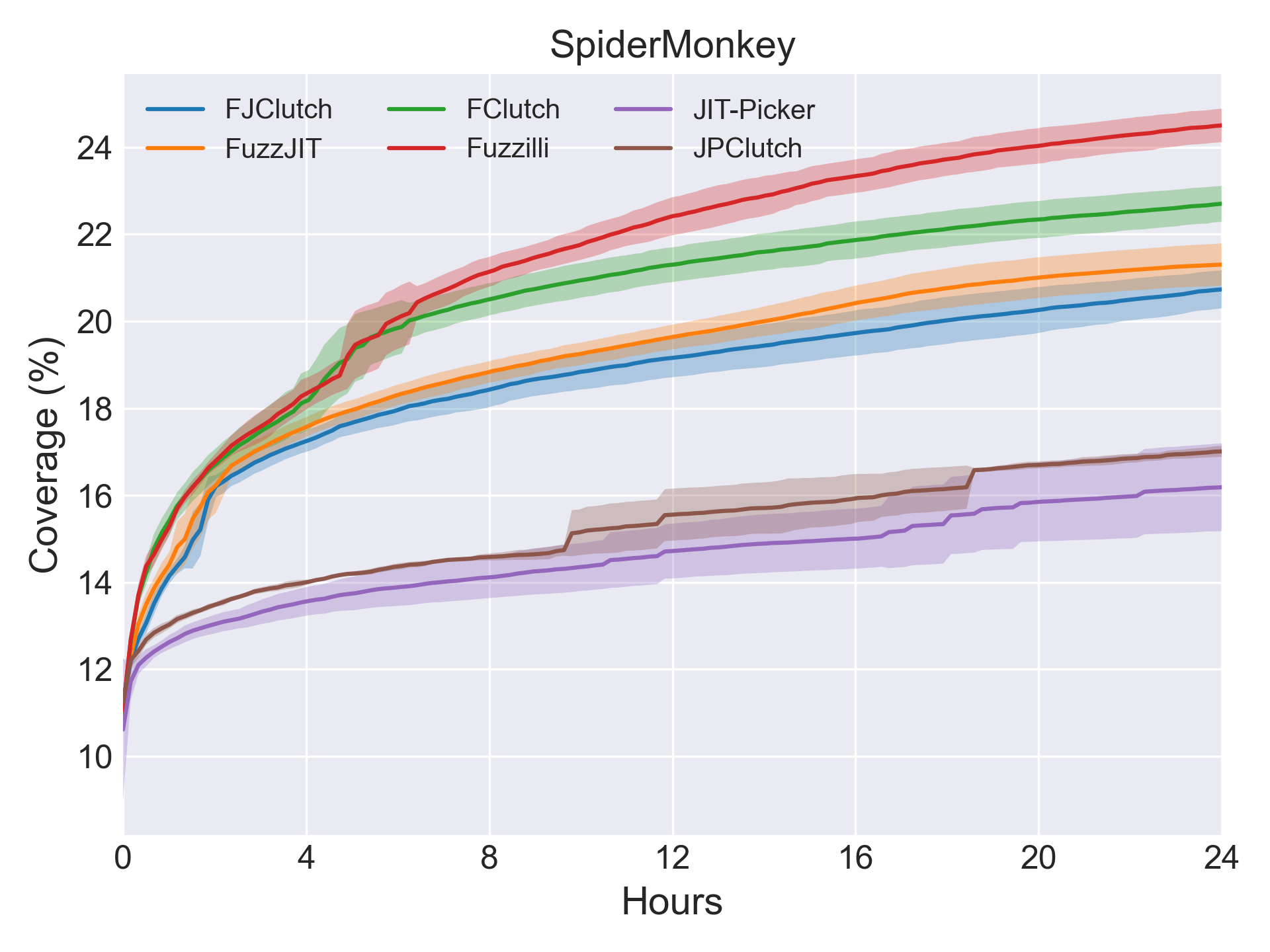}
    \label{sm_cov}}
\subfigure[]{
    \includegraphics[width=0.33\linewidth]{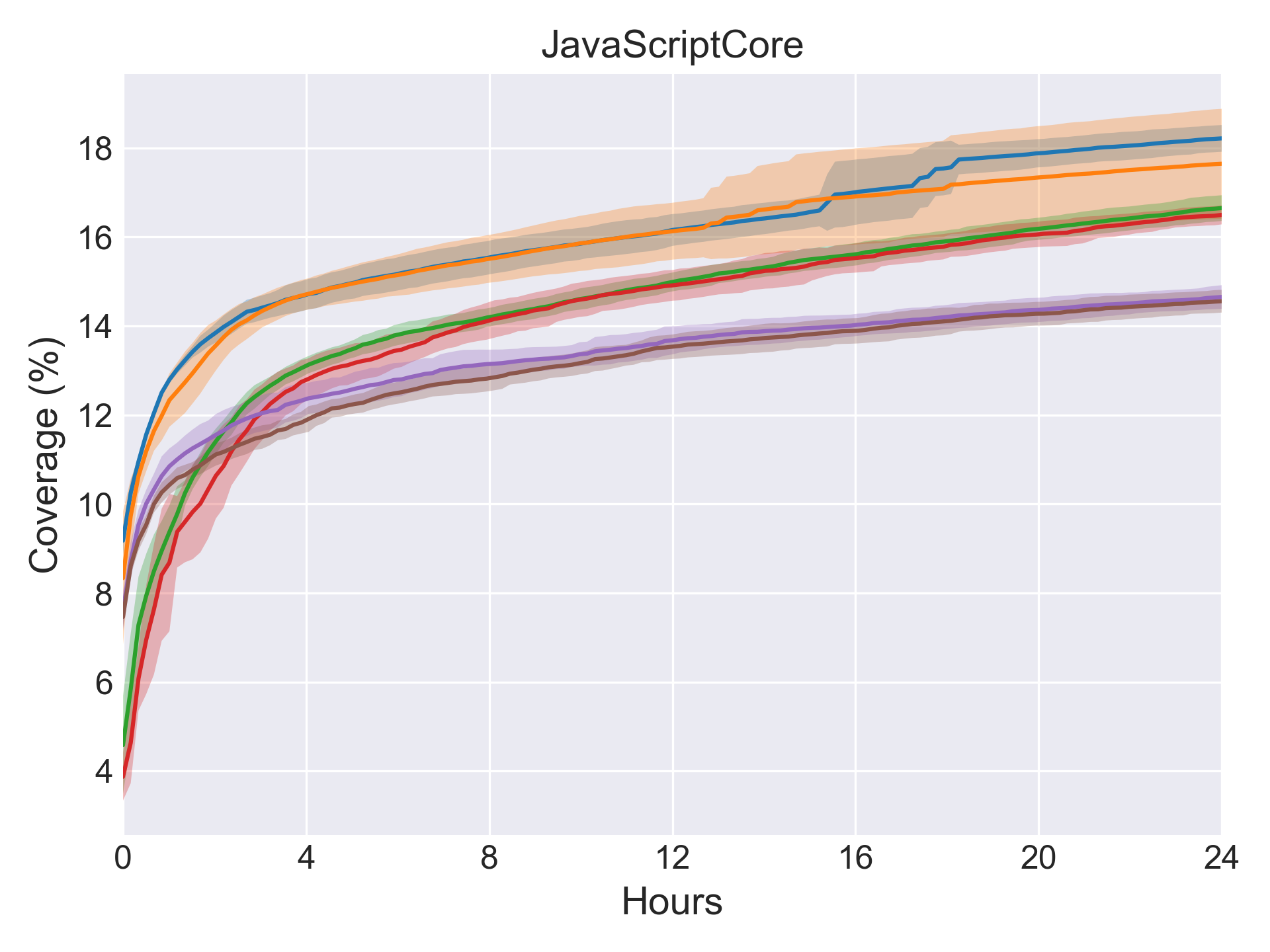}
    \label{jsc_cov}}

\caption{Branch coverage (five run average and standard deviation) from 24h fuzzing of \js\ engines.
}\label{fig:coverage}
\end{figure}

Coverage is often a primary metric for evaluation of fuzzers. 
We show the total branch coverage in Table~\ref{tab:coverage}, and for completeness we report the coverage over time in Figure~\ref{fig:coverage}.

We observe a limited and mixed effect on overall coverage by adopting \clutch, as shown by the average and relative change percentages, and the small values of $\hat{A}_{12}$ in four instances. 
Still, 4 of the \clutch\ based models do outperform the fuzzer baselines in terms of coverage achieved.  
In fact, on average, across all combinations of fuzzer and \js\ engine \clutch\ improves the coverage by an average of 2.3\%.

Given the lower number of test cases that \clutch\ variants execute, this aggregate performance is remarkable.
\clutch\ leverages its ability to select the location for mutations, improving the efficiency of testing.
%
To perform a more legitimate comparison on this efficiency we compare the coverage relative to the number of test cases executed in Section~\ref{eval:ftc:efficiency}.

\subsection{Comparison to \clite}\label{eval:accub}


%


We have shown how \clutch\ outperforms equivalent fuzzing baselines.
%
Compared with random selection, \clutch\ learns from its experience to improve efficiency, and a targeted testing strategy. 
We designed \clutch\ to handle the many complex problems that come with selecting a location to perform mutations, as outlined in Sections~\ref{fuzz:challenge} and~\ref{clutch}. 
Yet, it remains to be seen if alternative bandit architectures could learn in such a setting.

Thus we introduce \clite, an alternative bandit model based on the Adaptive Contextual Combinatorial Upper Confidence Bound (ACC-UCB)~\cite{nika_contextual_2020} algorithm. 
While ACC-UCB was originally used for solving `dynamic resource allocation problems' using a tree based structure, we have modified it to work within \clutch, by providing it with the ability to increase the search space over time (\ie\ the number of instructions seen from test cases over time).
\clite\ can then handle the context of selecting instructions from the IR, and the variable number of arms that are seen over the course of fuzzing.
As a result we can compare \clutch\ to another contextual bandit for selecting locations to mutate.
We compare \clutch\ and \clite\ for their efficiency of branches found per execution, reporting the results in Table~\ref{tab:bandit_eff}. 

We can see that \clite\ learns to select appropriate locations for mutations, as evidenced by the comparable performance in efficiency compared to \clutch.
In \clite s best case it increases efficiency by 32.2\%, and at worst reduces efficiency by 25.1\%. 
This result shows that even alternative bandit algorithms are a capable of learning to optimise mutations when fuzzing.
While \clite\ has been able to learn, we still see the superior performance from \clutch, resulting in greater branch coverage per execution in the majority of engines, increasing the aggregate efficiency by 30.81\%, and 3.42\% on average.


\begin{table*}
\centering
\caption{Branch coverage per generated test case (five run average) from 24h fuzzing of \clutch\ and \clite.}
\label{tab:bandit_eff}
{\scriptsize
\begin{tabular}{@{}cr|cc|cc|cc@{}}
\toprule
Engine & Metric & \fjc & \fja & \fzc & \fza & \jpc & \jpa \\ \midrule
\multirow{4}{*}{V8} & Average & \textbf{0.627} & 0.620 & \textbf{0.387} & 0.382 & \textbf{0.585} & 0.438 \\
 & Relative Change & \multicolumn{2}{c|}{-1.13\%} & \multicolumn{2}{c|}{-1.24\%} & \multicolumn{2}{c}{-25.1\%} \\
 & $\hat{A}_{12}$ & \multicolumn{2}{c|}{0.560} & \multicolumn{2}{c|}{0.480} & \multicolumn{2}{c}{0.880} \\
 & $p$ & \multicolumn{2}{c|}{0.841} & \multicolumn{2}{c|}{1.000} & \multicolumn{2}{c}{0.056} \\\midrule
\multirow{4}{*}{SpiderMonkey} & Average & \textbf{0.067} & 0.058 & 0.077 & \textbf{0.088} & 0.081 & \textbf{0.103} \\
 & Relative Change & \multicolumn{2}{c|}{-13.4\%} & \multicolumn{2}{c|}{13.52\%} & \multicolumn{2}{c}{26.06\%} \\
 & $\hat{A}_{12}$ & \multicolumn{2}{c|}{1.000} & \multicolumn{2}{c|}{0} & \multicolumn{2}{c}{0.111} \\
 & $p$ & \multicolumn{2}{c|}{0.008} & \multicolumn{2}{c|}{0.008} & \multicolumn{2}{c}{0.143} \\ \midrule
\multirow{4}{*}{JavaScriptCore} & Average & 0.508 & \textbf{0.671} & \textbf{0.240} & 0.235 & 0.358 & \textbf{0.363} \\
 & Relative Change & \multicolumn{2}{c|}{32.20\%} & \multicolumn{2}{c|}{-2.07\%} & \multicolumn{2}{c}{1.968\%} \\
 & $\hat{A}_{12}$ & \multicolumn{2}{c|}{0.000} & \multicolumn{2}{c|}{0.680} & \multicolumn{2}{c}{0.500} \\
 & $p$ & \multicolumn{2}{c|}{0.008} & \multicolumn{2}{c|}{0.421} & \multicolumn{2}{c}{0.571} \\ \bottomrule
 \end{tabular}}
\end{table*}

\section{Additional Combinatorial Bandit Evaluation}\label{app:furtherbandit}

\begin{figure*}
\centering
\subfigure[]{
    \includegraphics[width=0.4\linewidth]{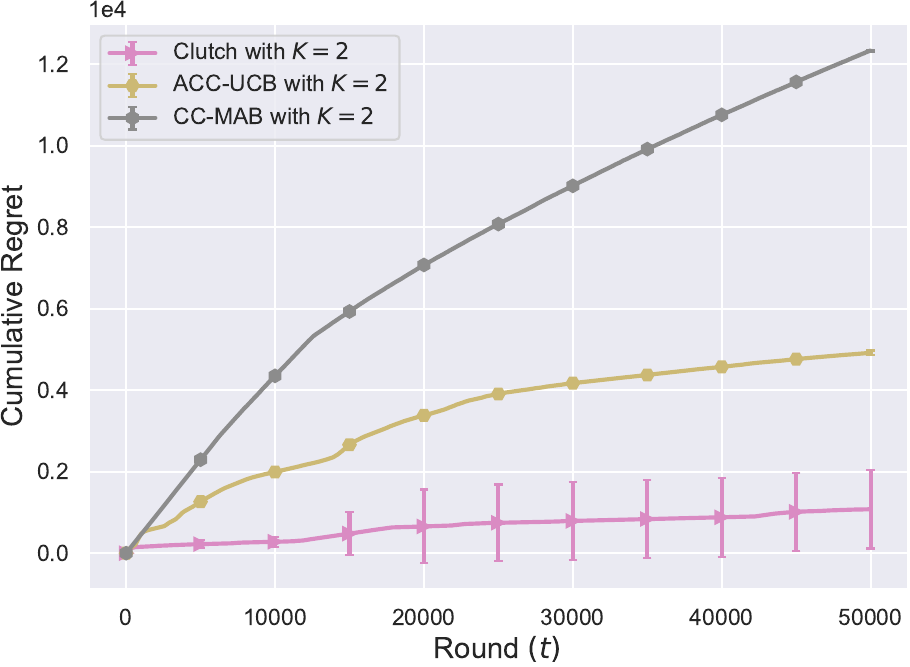}
    \label{gorilla_regret_2}}
\subfigure[]{
    \includegraphics[width=0.4\linewidth]{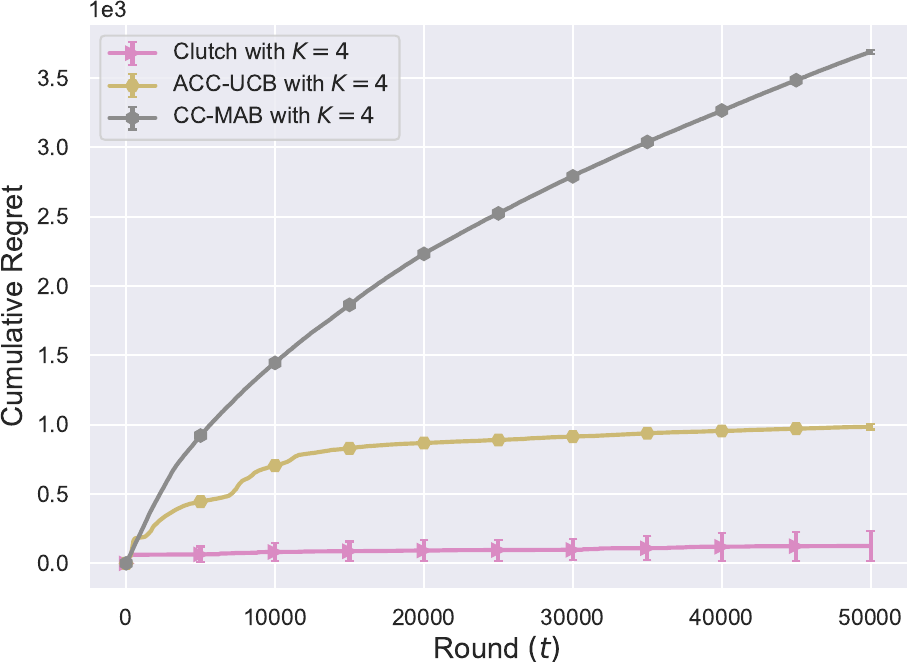}
    \label{gorilla_regret_4}}  
\caption{Cumulative regret of volatile combinatorial contextual bandits.}\label{fig:gorilla_regret}
\end{figure*}

We include here the plots pertaining to the experiments detailed in Section~\ref{sec:bandit_comp} and Table~\ref{tab:bandit_comp}. Figure~\ref{fig:gorilla_regret} contains the regret of bandits in the volatile setting of the Gollawa dataset used by Nika \etal~\cite{nika_contextual_2020}. 
Table~\ref{tab:bandit_sig} includes the statistical testing values related to effect size (significant when $\hat{A}_{12}\geq 0.71$) and statistical significance (significant when $p<0.1$).

\begin{figure*}[t]
\centering
\subfigure[]{
    \includegraphics[width=0.31\linewidth]{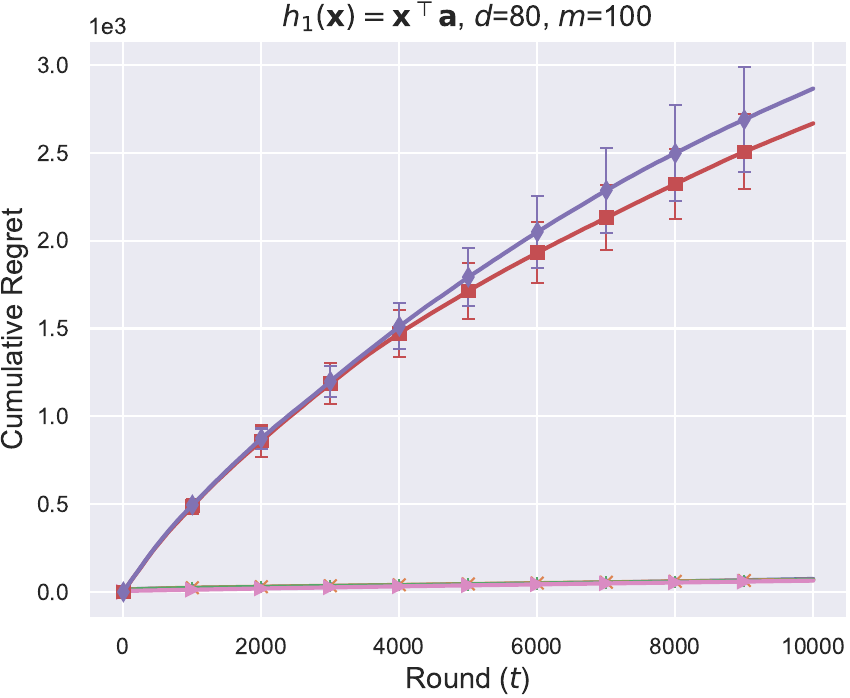}
    \label{h1}}  
\subfigure[]{
    \includegraphics[width=0.31\linewidth]{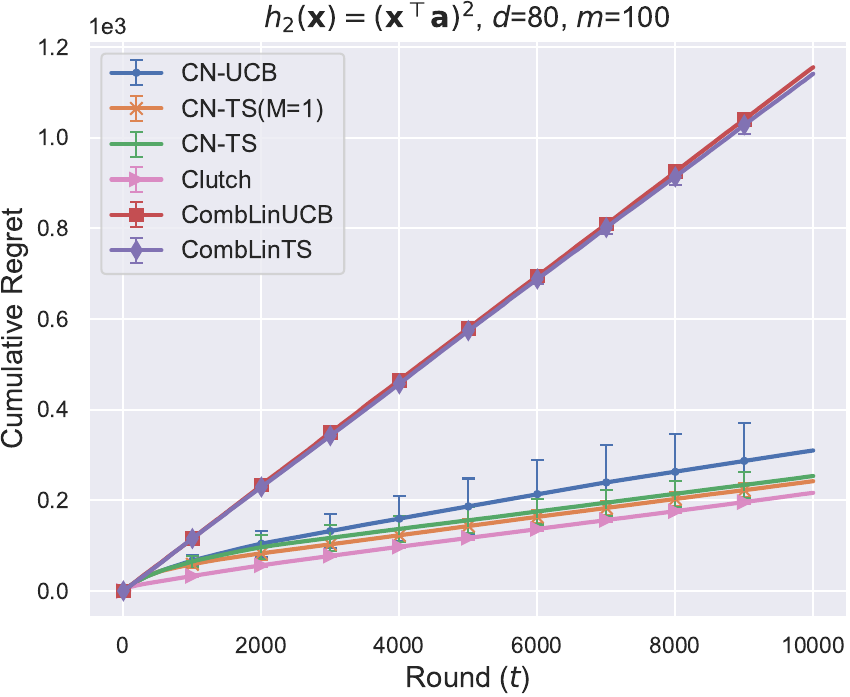}
    \label{h2}}  
\subfigure[]{
    \includegraphics[width=0.31\linewidth]{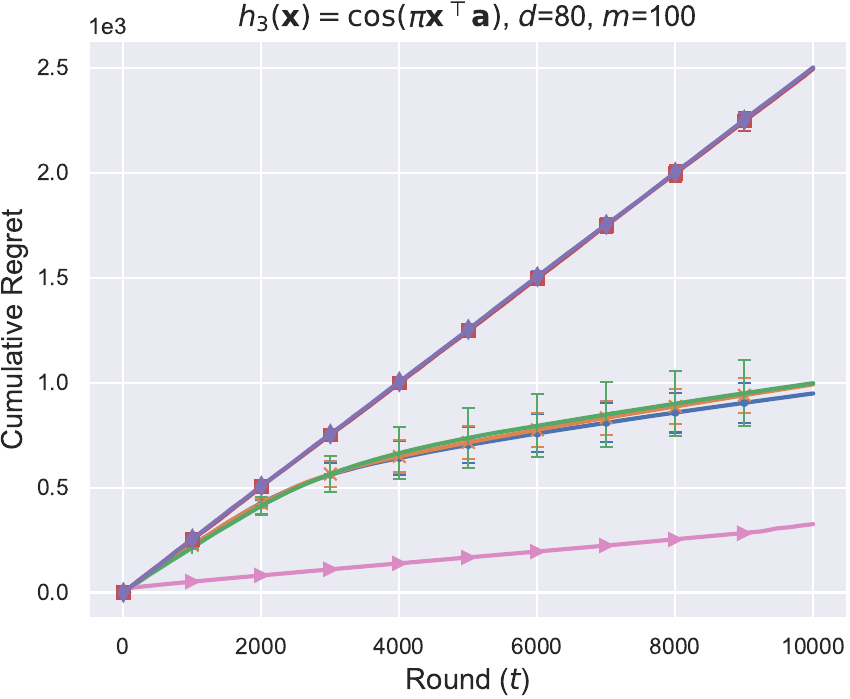}
    \label{h3}}

\subfigure[]{
    \includegraphics[width=0.31\linewidth]{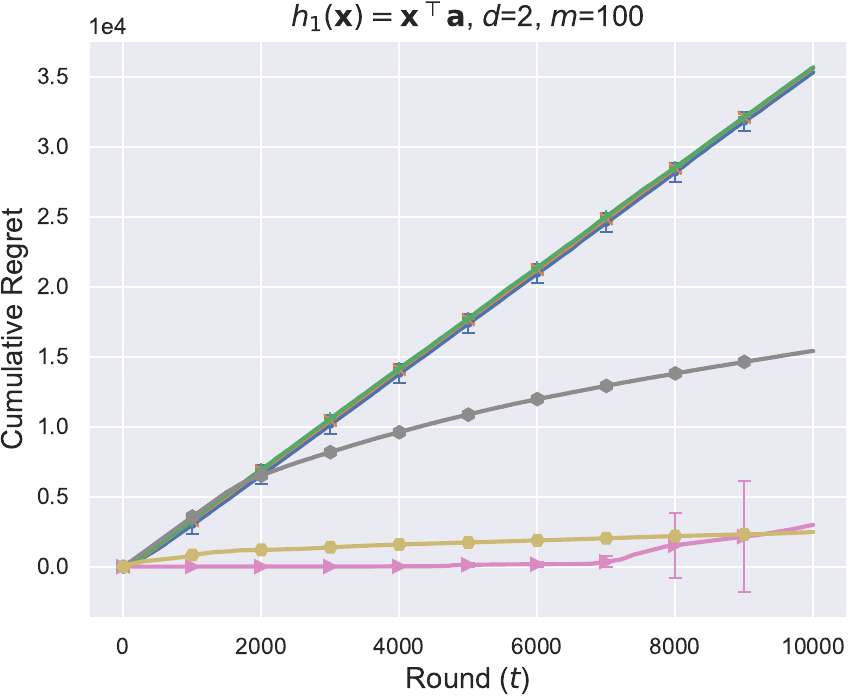}
    \label{h1_2}}  
\subfigure[]{
    \includegraphics[width=0.31\linewidth]{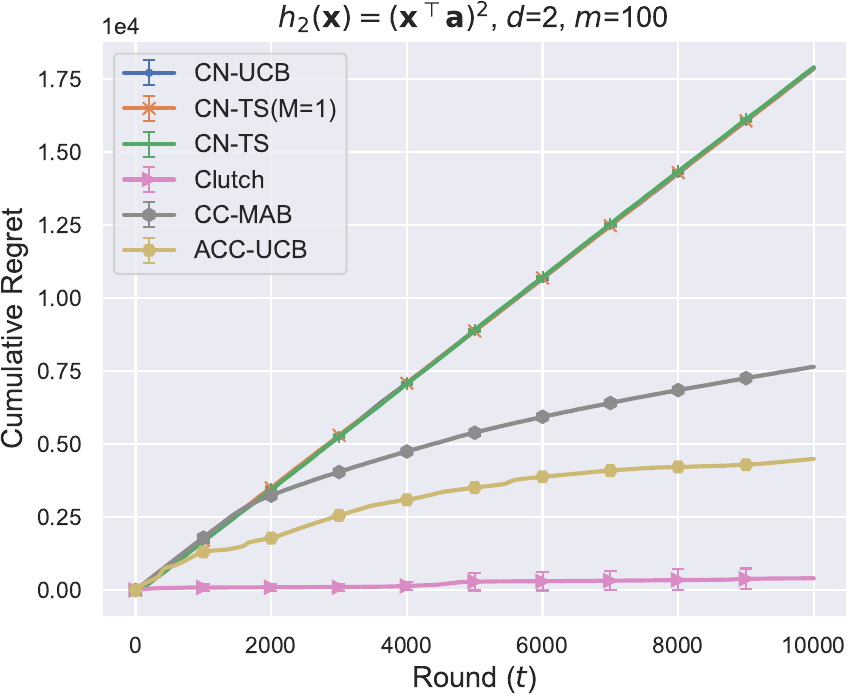}
    \label{h2_2}}  
\subfigure[]{
    \includegraphics[width=0.31\linewidth]{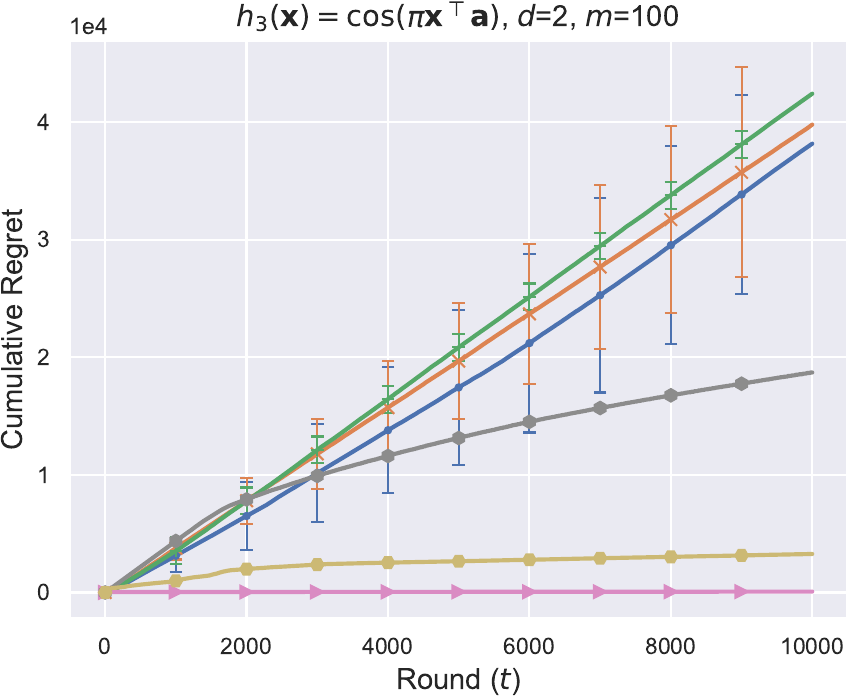}
    \label{h3_2}}
\caption{Cumulative regret of \clutch\ compared with deep and linear combinatorial contextual bandits. Top row: 80 features per arm ($d$) and 100 nodes per hidden layer ($m$). Bottom row:  80 features per arm ($d$) and 100 nodes per hidden layer ($m$).}\label{fig:unknown_2}
\end{figure*}

For the combinatorial setting (without volatile arms) we include plots of regret when there are 80 observable features per arm and two features per arm (Figure~\ref{fig:unknown_2}).
As discussed in Section~\ref{sec:bandit_comp} the volatile bandits require exactly two features per arm in order run their respective bandit algorithms. 
In order to compare across all the bandits in an equal setting we set the number of features per arm in the combinatorial setting to 2.
The results from this experiment is presented in Table~\ref{tab:dim=2} and Figure~\ref{fig:unknown_2}.
In two of the three reward functions ($h_2$ and $h_3$) \clutch\ achieves the lowest regret of any bandit. 
However, in reward $h_1$, ACC-UCB outperforms \clutch\ by 16.8\%. This is due to an anomaly in one of the instances of training \clutch, where it performs 6.5 times worse causing the standard deviation to become significantly larger than in other instances. This behaviour is observed in the last 3000 timesteps in Figure~\ref{h1_2}, where the standard deviation increases.



\begin{table*}[]
\centering
\caption{Statistical significance of bandit regret performance compared to \clutch, using Vargha and Delaney’s $\hat{A}_{12}$ test and Mann-Whitney U test ($p$ values).}
\label{tab:bandit_sig}
\resizebox{\textwidth}{!}{
\begin{tabular}{@{}ccccccccccc@{}}
\toprule
\multirow{2}{*}{Setting} & \multirow{2}{*}{Volatile} & \multirow{2}{*}{Selected Arms} & \multirow{2}{*}{Features} & \multicolumn{7}{c}{$\hat{A}_{12}$, $p$} \\
 &  &  &  & ACC-UCB & CC-UCB & CombLinTS & CombLinUCB & CN-UCB & CN-TS & CN-TS(M=1) \\ \midrule
\multirow{2}{*}{Golwalla Dataset} & \multirow{2}{*}{\checkmark} & 2 & \multirow{2}{*}{2} & 1.0, 0.0002 & 1.0, 0.0002 & - & - & - & - & - \\
 &  & 4 &  & 1.0, 0.0002 & 1.0, 0.0002 & - & - & - & - & - \\
\multirow{2}{*}{$R_{1}(\mathbf{x}) = \mathbf{x}^{\top}\mathbf{a}$} &  & 4 & 2 & 0.8, 0.03 & 0.9, 0.03 & 0.78, 0.04 & 0.9, 0.03 & 1.0, 0.0002 & 1.0, 0.0002 & 1.0, 0.0002 \\
 &  &  & 80 & - & - & 1.0, 0.0002 & 1.0, 0.0002 & 0.98, 0.0003 & 0.81, 0.02 & 0.86, 0.007 \\
\multirow{2}{*}{$R_{2}(\mathbf{x}) = (\mathbf{x}^{\top}\mathbf{a})^{2}$} &  & \multirow{2}{*}{4} & 2 & 1.0, 0.0002 & 1.0, 0.0002 & 1.0, 0.0002 & 1.0, 0.0002 & 1.0, 0.0002 & 1.0, 0.0002 & 1.0, 0.0002 \\
 &  &  & 80 & - & - & 1.0, 0.0002 & 1.0, 0.0002 & 1.0, 0.0002 & 0.96, 0.0006 & 1.0, 0.0002 \\
\multirow{2}{*}{$R_{3}(\mathbf{x}) = \cos(\pi \mathbf{x}^{\top}\mathbf{a})$} &  & \multirow{2}{*}{4} & 2 & 1.0, 0.0002 & 1.0, 0.0002 & 1.0, 0.0002 & 1.0, 0.0002 & 1.0, 0.0002 & 1.0, 0.0002 & 1.0, 0.0002 \\
 &  &  & 80 & - & - & 1.0, 0.0002 & 1.0, 0.0002 & 1.0, 0.0002 & 1.0, 0.0002 & 1.0, 0.0002 \\ \bottomrule
\end{tabular}}
\end{table*}

\begin{table}[t]
\caption{Total regret of combinatorial bandits in combinatorial bandit settings with two features per arm.}\label{tab:dim=2}
\resizebox{\textwidth}{!}{
\begin{tabular}{@{}ccccccccc@{}}
\toprule
\multirow{2}{*}{Setting} & \multicolumn{8}{c}{Total Regret} \\
 & ACC-UCB & CC-UCB & CombLinTS & CombLinUCB & CN-UCB & CN-TS & CN-TS(M=1) & \clutch\ \\ \midrule
$R_{1}(\mathbf{x}) = \mathbf{x}^{\top}\mathbf{a}$ & \textbf{\begin{tabular}[c]{@{}c@{}}2485.14 \\ ($\pm$22.22)\end{tabular}} & \begin{tabular}[c]{@{}c@{}}15405.27 \\ ($\pm$3.71)\end{tabular} & \begin{tabular}[c]{@{}c@{}}3400.14 \\ ($\pm$945.67)\end{tabular} & \begin{tabular}[c]{@{}c@{}}6470.49 \\ ($\pm$1061.56)\end{tabular} & \begin{tabular}[c]{@{}c@{}}35312.04 \\ ($\pm$667.05)\end{tabular} & \begin{tabular}[c]{@{}c@{}}35669.80 \\ ($\pm$16.68)\end{tabular} & \begin{tabular}[c]{@{}c@{}}35604.24 \\ ($\pm$233.58)\end{tabular} & \begin{tabular}[c]{@{}c@{}}2988.15 \\ ($\pm$5641.32)\end{tabular} \\
$R_{2}(\mathbf{x}) = (\mathbf{x}^{\top}\mathbf{a})^{2}$ & \begin{tabular}[c]{@{}c@{}}4490.68 \\ ($\pm$11.95)\end{tabular} & \begin{tabular}[c]{@{}c@{}}7645.91 \\ ($\pm$3.38)\end{tabular} & \begin{tabular}[c]{@{}c@{}}15851.85 \\ ($\pm$9413.66)\end{tabular} & \begin{tabular}[c]{@{}c@{}}9408.59 \\ ($\pm$4772.00)\end{tabular} & \begin{tabular}[c]{@{}c@{}}17862.91 \\ ($\pm$40.10)\end{tabular} & \begin{tabular}[c]{@{}c@{}}17905.96 \\ ($\pm$50.17)\end{tabular} & \begin{tabular}[c]{@{}c@{}}17882.48 \\ ($\pm$46.23)\end{tabular} & \textbf{\begin{tabular}[c]{@{}c@{}}402.20 \\ ($\pm$353.64)\end{tabular}} \\
$R_{3}(\mathbf{x}) = \cos(\pi \mathbf{x}^{\top}\mathbf{a})$ & \begin{tabular}[c]{@{}c@{}}3273.88 \\ ($\pm$17.58)\end{tabular} & \begin{tabular}[c]{@{}c@{}}18699.54 \\ ($\pm$8.21)\end{tabular} & \begin{tabular}[c]{@{}c@{}}43407.08 \\ ($\pm$193.73)\end{tabular} & \begin{tabular}[c]{@{}c@{}}43171.84 \\ ($\pm$61.45)\end{tabular} & \begin{tabular}[c]{@{}c@{}}38149.58 \\ ($\pm$8426.18)\end{tabular} & \begin{tabular}[c]{@{}c@{}}42397.60 \\ ($\pm$1143.33)\end{tabular} & \begin{tabular}[c]{@{}c@{}}39768.50 \\ ($\pm$9970.75)\end{tabular} & \textbf{\begin{tabular}[c]{@{}c@{}}77.05 \\ ($\pm$30.18)\end{tabular}} \\ \bottomrule
\end{tabular}}
\end{table}

\end{document}